\documentclass[10pt,twocolumn,letterpaper]{article}

\def\paperID{477} 
\def\printOption{arxiv} 

\def\printFinal{final}
\def\printReview{review}
\def\printArxiv{arxiv}

\ifx \printOption \printFinal
	\usepackage{wacv}  	
\else
	\ifx \printOption \printReview
		\usepackage[review,algorithms]{wacv}
	\else
		\usepackage[pagenumbers]{wacv} 
	\fi
\fi

\usepackage[accsupp]{axessibility} 

\usepackage{import}

\usepackage[utf8]{inputenc} 
\usepackage[T1]{fontenc}    

\usepackage{amsmath}
\usepackage{amsthm}
\usepackage{mathtools}
\usepackage{commath}
\usepackage{mathrsfs}
\usepackage{amssymb}
\usepackage{amsfonts}       
\usepackage{bbm}
\usepackage{bm}
\usepackage{fancyhdr}
\usepackage{cancel}
\usepackage{nicefrac}       
\usepackage{microtype}      
\usepackage{fontawesome } 

\usepackage{algorithm}
\usepackage[noend]{algpseudocode}
\usepackage{eqparbox}

\usepackage{wrapfig}

\usepackage[pdftex]{graphicx}
\usepackage{epsfig}
\usepackage[utf8]{inputenc}
\usepackage{tabularx}
\usepackage{booktabs, multicol, multirow}	
\usepackage[table, xcdraw]{xcolor}

\usepackage{caption}
\usepackage{subcaption}

\usepackage{appendix}

\usepackage{cite}
\usepackage{url}            

\usepackage{comment}

\usepackage[pagebackref,breaklinks,colorlinks]{hyperref}

\usepackage[capitalize]{cleveref}
\crefname{section}{\S}{\S\S}
\Crefname{section}{Section}{Sections}
\Crefname{table}{Table}{Tables}
\crefname{table}{Tab.}{Tabs.}

\def\wacvPaperID{\paperID} 
\def\confName{WACV}
\def\confYear{2024}

\ifx \printOption \printFinal
	\def\codeurl{https://github.com/yetigurbuz/ccp-dml} 
        
\else
       
	\ifx \printOption \printReview
		\def\codeurl{https://drive.google.com/drive/folders/1KY6ptbgbdOqzYWnrU7JR4zgLAigxC5r4?usp=sharing} 
	\else
		\def\codeurl{https://github.com/yetigurbuz/ccp-dml} 
	\fi
\fi

\DeclareMathOperator*{\argmin}{arg\,min}
\DeclareMathOperator*{\argmax}{arg\,max}

\newcommand{\E}[2]{\ensuremath{\operatorname{\mathbb{E}}_{#2}[ {#1} ]}}

\newcommand{\ind}[1]{\ensuremath{\mathbbm{1}({#1})}}
\newcommand{\Real}{I\!\!R} 

\DeclareMathSymbol{\shortminus}{\mathbin}{AMSa}{"39}

\newcommand{\sxtimes}{\mathsf{x}\mskip1mu}

\usepackage{xspace}
\makeatletter
\DeclareRobustCommand\onedot{\futurelet\@let@token\@onedot}
\def\@onedot{\ifx\@let@token.\else.\null\fi\xspace}

\def\eg{\emph{e.g}\onedot} 
\def\ie{\emph{i.e}\onedot} \def\Ie{\emph{I.e}\onedot}

\makeatother

\theoremstyle{plain}
\newtheorem{theorem}{Theorem}[section]
\newtheorem{proposition}[theorem]{Proposition}
\newtheorem{lemma}[theorem]{Lemma}
\newtheorem{corollary}[theorem]{Corollary}
\theoremstyle{definition}

\theoremstyle{remark}

\newtheorem{innercustomprop}{Proposition}
\newenvironment{customprop}[1]
  {\renewcommand\theinnercustomprop{#1}\innercustomprop}
  {\endinnercustomprop}

\newtheorem{innercustomlemma}{Lemma}
\newenvironment{customlemma}[1]
  {\renewcommand\theinnercustomlemma{#1}\innercustomlemma}
  {\endinnercustomlemma}

\newtheorem{innercustomcorr}{Corollary}
\newenvironment{customcorr}[1]
  {\renewcommand\theinnercustomcorr{#1}\innercustomcorr}
  {\endinnercustomcorr}

\numberwithin{equation}{section}

\makeatletter
\def\BState{\State\hskip-\ALG@thistlm}
\makeatother

\begin{document}

\title{Deep Metric Learning with Chance Constraints}

\def\yeti{Yeti~Z.~G\"{u}rb\"{u}z}
\def\ogul{O\u{g}ul~Can}
\def\ada{Ada~G\"{o}rg\"{u}n}
\def\aydin{A.~Ayd{\i}n~Alatan}

\renewcommand\footnotemark{}

\def\instmetu{OGAM and METU}
\def\addrssmetu{METU, TR}

\def\instmetadialog{MetaDialog}
\def\instcerebrate{Cerebrate AI}
\def\addrssmetadialog{TU Berlin, DE}

\def\instuni{Faculty of Elect. Elec. Eng. and Computer Science}
\def\addrssuni{Somewhere in Europe}

\def\instapple{Intel Labs}

\def\emailyeti{yeti@metu.edu.tr}
\def\emailogul{ogul.can}
\def\emailala{alatan@metu.edu.tr}
\def\emailada{ada.gorgun@metu.edu.tr}


\newcommand{\affinfo}{}

\author{\yeti$^{\dagger}$ \qquad \ogul$^{\dagger\dagger}$ \qquad \aydin$^{\ddagger}$\\
\instmetu\\
\addrssmetu\\
{\tt\small\{$^{\dagger}$yeti, $^{\dagger\dagger}$ogul.can, $^{\ddagger}$alatan\}@metu.edu.tr}
}

\author{\begin{tabular}{ccc}
\yeti\textsuperscript{$\dagger$,1 }\thanks{\textsuperscript{$\dagger$}Affiliated with OGAM-METU during the research.} & \ogul\textsuperscript{1, 2} & \aydin\textsuperscript{3} \\
     \textsuperscript{1}\instmetadialog & \textsuperscript{2}\instcerebrate & \textsuperscript{3}\instmetu \\
\end{tabular}
}

\maketitle

\begin{abstract}
\label{sec:abstract}
Deep metric learning (DML) aims to minimize empirical expected loss of the pairwise intra-/inter- class proximity violations in the embedding space. We relate DML to feasibility problem of finite chance constraints. We show that minimizer of proxy-based DML satisfies certain chance constraints, and that the worst case generalization performance of the proxy-based methods can be characterized by the radius of the smallest ball around a class proxy to cover the entire domain of the corresponding class samples, suggesting multiple proxies per class helps performance. To provide a scalable algorithm as well as exploiting more proxies, we consider the chance constraints implied by the minimizers of proxy-based DML instances and reformulate DML as finding a feasible point in intersection of such constraints, resulting in a problem to be approximately solved by iterative projections. Simply put, we repeatedly train a regularized proxy-based loss and re-initialize the proxies with the embeddings of the deliberately selected new samples. We applied our method with 4 well-accepted DML losses and show the effectiveness with extensive evaluations on 4 popular DML benchmarks. Code is available at: \href{\codeurl}{CCP-DML Framework}
\end{abstract}


\ifx \printOption \printArxiv
\fancypagestyle{firststyle}
{
   \fancyhead{}
   \lhead{Accepted as a conference paper at WACV 2024}
}
\thispagestyle{firststyle}
\def\refappendix{\hyperref[sec:all_proofs]{appendix}}
\def\refmagnified{\textsuperscript{\protect\hyperlink{figmagnified}{\faSearch}}} 
\def\refexpdetail{\cref{sec:setup_supp} }
\def\refmlrctab{\Cref{tab:all_results_bigger_cars_cub,tab:all_results_bigger_sop_inshop} in supplementary material }
\def\refablation{\cref{sec:sup_ablation}}

\else
\def\refappendix{appendix \cite{gurbuz_appendix}}
\def\refmagnified{\cite{gurbuz_appendix}} 
\def\refexpdetail{\cite[\S~2.1]{gurbuz_appendix} }
\def\refmlrctab{in supplementary material \cite[Tabs.~1 and 2]{gurbuz_appendix} }
\def\refablation{\cite[\S~1.2]{gurbuz_appendix}}

\fi


\section{Introduction}
\label{sec:introduction}
Deep metric learning (DML) poses distance metric problem as learning the parameters of an embedding function so that the semantically similar samples are embedded to the small vicinity in the representation space as the dissimilar ones are placed relatively apart in the Euclidean sense. The typical embedding function is implemented as convolutional neural networks (CNN) for visual tasks and the parameters are learned through minimizing the empirical expected loss with possibly deliberately selected mini-batch gradient updates \cite{roth2020revisiting, musgrave2020metric}. The loss terms in the empirical loss penalize violations of the desired intra- and inter-class proximity constrains. Large-scale problems (in terms of $\#classes$) suffer from the noisy estimation of the expected loss with mini-batches \cite{schroff2015facenet, wang2020cross, musgrave2020metric}. Recently, augmenting the mini-batches with virtual embeddings called \textit{proxies} is shown to better approximate empirical loss in large-scale problems \cite{wang2020cross, musgrave2020metric} owing to pseudo-global consideration of the dataset during loss computation. These advances raise a critical question: \textit{"How does increasing proxies help?"} which is partially addressed empirically with the methods exploiting multiple proxies per class \cite{Qian_2019_ICCV, wang2020cross, zhu2020fewer}. 


\begin{figure}[!t]
\centering
  \centerline{\includegraphics[width=1.\linewidth,keepaspectratio]{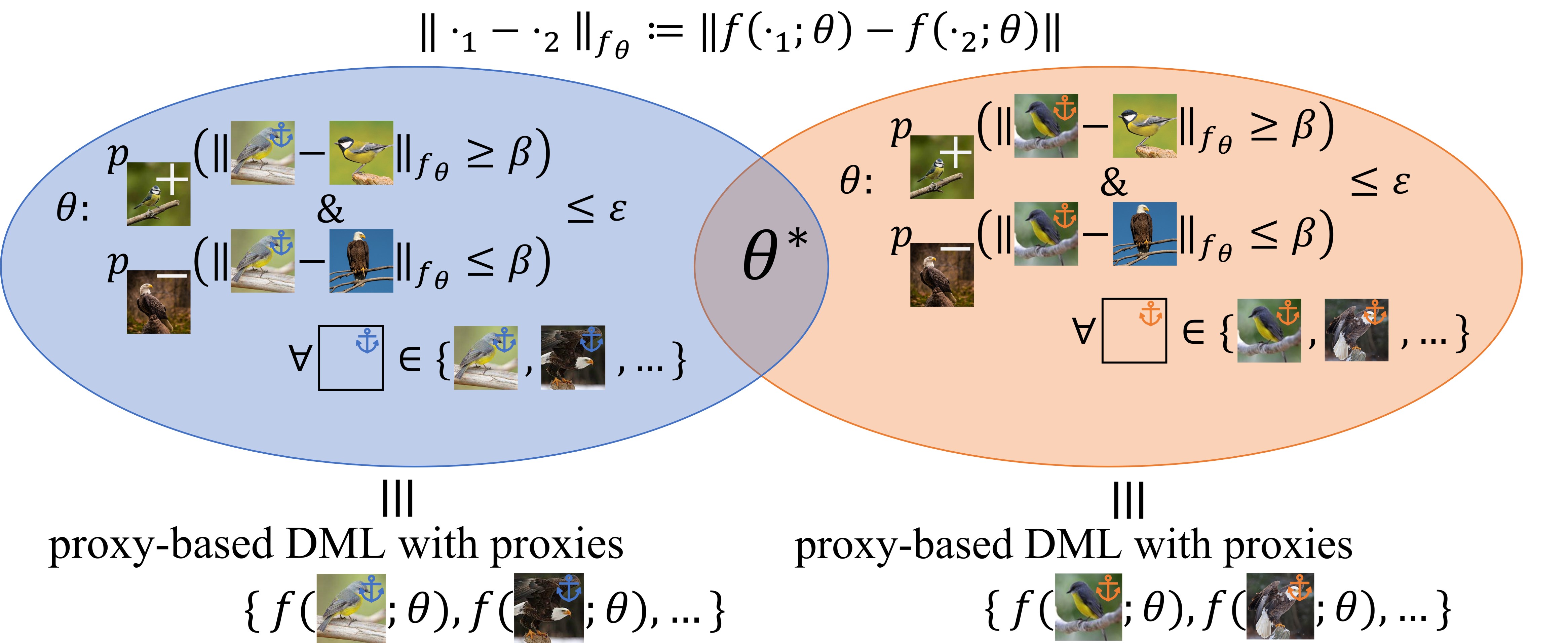}}
  \caption{Simple illustration of our chance constrained DML formulation over two sets where each set's elements are the embedding function parameters $\theta$ that yield an embedding space in which the distances to anchor samples assess the semantic dissimilarity with high probability (\ie, satisfying chance constraints). We consider the parameters of the desired embedding function $f(\cdot;\theta^\ast)$ to lie in the intersection of such sets. We show that to each such set, there corresponds a proxy-based DML solution. Hence, we solve DML as a set intersection problem via solving multiple proxy-based DML problems.}
	\label{fig:summary}
\end{figure}
%
%
%
%
%
%
%


Characterizing generalization performance of proxy-based DML can be a decisive step towards theoretically addressing that question. To this end, we approach DML differently by posing it as a feasibility problem. In particular, we consider a chance constraint \cite{birge2011introduction} for the desired embedding function and relate it to the typical expected loss of DML. Using such a relation, we provide an upper bound to the generalization error of proxy-based DML. Aligned with the literature, the form of the bound suggests possible room for the improvement on the generalization performance if more and diverse proxies are considered per class. However, straightforward increase of the proxies may not help; since, $i)$ proxies of the same class tend to merge \cite{Qian_2019_ICCV} and $ii)$ memory is prohibitive to arbitrarily increase the proxies. 

To alleviate these limitations, we relate the minimizer of the proxy-based DML to a feasible point of some chance constraints, and reformulate DML as finding a point at the intersection of the sets that the proxies imply. We provide a scalable algorithm using iterative projections to the individual sets to solve the problem. Each projection corresponds to a regularized proxy-based DML. Hence, we inherently increase the number of diverse proxies included in the problem. We empirically study the implications of our formulation and show its effectiveness by applying our method on 4 DML losses and testing on 4 DML benchmarks. Results show consistent and state-of-the-art (SOTA) performance in improving the baselines.

\section{Related Work}
\label{sec:review}
We discuss the works which are most related to ours. Briefly, our contributions include that $i)$ we reformulate DML as a chance constrained feasibility problem, $ii)$ we characterize generalization of proxy-based DML by expanding on the discussions of the works studying generalization bounds, $iii)$ we write the feasibility problem, \ie, DML, as a set intersection problem to be solved by iterative projections using proxy-based DML, and $iv)$ we effectively utilize arbitrary number of  proxies per class.


\textbf{DML.} Primary momentum in DML includes $i)$ tailoring the loss terms \cite{musgrave2020metric} to impose the desired intra- and inter-class proximity constraints in the representation space, $ii)$ pair mining \cite[and the references therein]{roth2020revisiting} to increase diversity in the loss computation or to reduce noise \cite{liu2021noise}, and $iii)$ synthesizing informative samples with generative models \cite{duan2018deep, zhao2018adversarial, lin2018deep, Zheng_2019_CVPR} or via interpolation \cite{ko2020embedding, proxysynthesis, venkataramanan2022it}. To improve embedding quality, detaching class-discriminative and class-shared features \cite{lin2018deep, Roth_2019_ICCV, gurbuz2023generalized}, intra-batch feature aggregation \cite{intrabatch,lim2022hypergraph}, ranking surrogates \cite{patel2022recall}, and further regularization terms \cite{Jacob_2019_ICCV, zhang2020deepSEC, kim2021multi, roth2022non} are utilized. Going beyond of a single model and loss, ensemble \cite{xuan2018deep, kim2018attention, Sanakoyeu_2019_CVPR, zheng2021deep2, zheng2021deep} and multi-task based approaches \cite{milbich2020diva, roth2021s2sd, ebrahimpour2022multi} are also used. Different to them, we approach DML from a unique perspective, redefining it as a set intersection problem with chance constraints.


\textbf{Ranking losses in DML.} Typical DML objective enforces distance ranking constraints among the samples in the embedding space via hinge losses penalizing ranking violations. The contrastive \cite{hadsell2006dimensionality,hu2014discriminative}, triplet \cite{weinberger2006distance, schroff2015facenet}, and generalized contrastive with margin \cite{wu2017sampling} losses are the simplest forms of the pairwise distance ranking based losses. Proceeding approaches utilize smoothed versions of these losses by replacing hinge loss with log-sum-exp \cite{oh2016deep,Wang_2019_CVPR_MS} or soft-max \cite{sohn2016improved, Yu_2019_ICCV} expressions, which inherently employ ranking among more samples via soft-batch-mining. Until very recently, log-sum-exp based ranking loss has been revamped with Bayesian perspective \cite{kan2022contrastive}. Likewise, we rediscover contrastive loss as a surrogate loss for our chance constrained DML formulation. Different to existing work, our formulation gives algorithmic implications to solve the DML as a set intersection problem which indeed can be solved efficiently with proxy-based DML.

\textbf{Proxy-based DML.} Proxy-based methods consider augmenting the mini-batch with more samples for less noisy estimate of the expected loss and circumvent the costly embedding computation to include more samples in the mini-batches. Proxies typically are vectors representing embeddings of the class centers \cite{proxynca, chen2018virtual, teh2020proxynca++, kim2020proxy} and are trained along with the embedding function parameters. Non-trainable proxies are also exploited in \cite{wang2020cross,ko2021learning} to gradually augment mini-batch with previously computed embeddings. In proxy-based DML, the pairwise distances are computed between the proxies and the mini-batch samples. Thus, pseudo-global dataset geometry is considered during loss computation. To better represent global geometry, multiple proxies per class are considered in \cite{rippel2016metric, Qian_2019_ICCV, zhu2020fewer} and a hierarchical
structure is imposed to proxies in \cite{yang2022hierarchical} where the former two \cite{rippel2016metric, Qian_2019_ICCV} build on improving R@1 (immediate neighbourhood) by fine-grained clustering of class samples to overlook intra-class variances. In our analysis, we also align with increasing the proxies. Our work differs in that $i)$ we build on reducing the probability of proximity violations (\ie, improving MAP@R) and $ii)$ we progressively increase the proxies by relating the proxy-based DML instances.



\textbf{DML as constrained optimization.} Pioneer metric learning approaches \cite{xing2003distance, weinberger2006distance} consider sample-driven proximity constraints to formulate the problem and exploit alternating projections to perform projected gradient ascent. Recently, sample-driven constraints are reconsidered in \cite{profs} for a reformulation of DML as a set intersection. Sharing the set intersection concept, we approach the problem with a different perspective using chance constraints rather than sample-driven constraints, which enables us to formally develop a method that does not suffer from the poor scalability of exploiting class representatives unlike the method proposed in \cite{profs} does. What makes our method unique is the theoretically sound way we connect the proxy-based DML and set intersection concepts to arbitrarily increase the number of class representatives exploited in the problem.

\textbf{Characterizing generalization bounds.} Notion of robustness in learning algorithms is studied in \cite{xu2012robustness} and generalization error bounds of several techniques are derived accordingly. This study is extended to metric learning setting in \cite{bellet2015robustnessML}. These works study the deviation between the expected loss and the empirical loss over the whole dataset. Differently in \cite{sener2018active}, deviation between two empirical losses, \textit{core-set loss}, is studied to characterize generalization loss when a subset of the training data is exploited. Generalization bound for metric learning is further studied in \cite{dong2020generalization,lei2021generalization} to analyze and suggest training strategies. Our work expands on the theories in the aforementioned works to characterize and improve generalization bound for proxy-based DML.

%
%
%
%
%
%

\section{Notation and Problem Definition}
\label{sec:notations}
In typical DML, we consider the set $\mathcal{Z}=\mathcal{X}\times\mathcal{Y}$ with elements $z=(x\in\mathcal{X},y\in\mathcal{Y})$ where $\mathcal{X}$ is a compact space and $\mathcal{Y}=\lbrace 1,\ldots, C\rbrace$ is a finite label set. We will use $x$ (or $y$) to denote data (or label) component of $z$. We have $p_\mathcal{Z}$, an unknown probability distribution over $\mathcal{Z}$. Indicator of the two samples, $z_i$ and $z_j$, belonging to the same class is denoted as $y_{ij}\in\lbrace \shortminus 1,1\rbrace$ where $y_{ij}=1$ if $y_i=y_j$. 

We are interested in finding the parameters $\theta$ of an embedding function $f(\cdot;\theta):\mathcal{X}{\xrightarrow{}}\Real^D$ so that the parametric distance, $\Vert x_i \shortminus x_j \Vert_{f_\theta} \coloneqq \Vert f(x_i;\theta)\shortminus f(x_j;\theta)\Vert_2$, is small only whenever $y_i=y_j$. For any pair $(z_i,z_j)\sim p_{\mathcal{Z}}$ and embedding function $f(\cdot;\theta)$, we associate a loss $\ell(z_i,z_j; \theta)$ penalizing proximity violations in the embedding image. We omit $f$ dependency in the $\ell$ notation for simplicity. We are to consider minimization of the expected loss:
\begin{equation}\label{eq:DML_true}
\theta^\ast=\argmin_\theta \E{\ell(z_i,z_j; \theta)}{z_i,z_j}
\end{equation}

In practice, we are given a dataset of $n$ instances sampled \textit{i.i.d.} from $\mathcal{Z}$ as $\lbrace z_i \rbrace_{i\in[n]}\sim p_\mathcal{Z}$  where \mbox{$[n]=\lbrace1,\ldots,n\rbrace$}, and an algorithm $\mathcal{A}_{s_1\sxtimes s_2}$ which outputs parameters $\theta$ minimizing empirical expected loss with a training error $e(\mathcal{A}_{s_1\sxtimes s_2})$ for a given set $\lbrace(z_i,z_j)\rbrace_{i,j\in s_1\sxtimes s_2}$ of pairs from the dataset, where $s_k=\lbrace s_k(l) \in [n]\rbrace_{l\in[n_k]}\subseteq[n]$ is a pool of indexes chosen from the dataset, $[n]$. In other words,
\begin{equation} \label{eq:empi}
\mathcal{A}_{s_1{\sxtimes}s_2} \coloneqq\argmin_\theta \tfrac{1}{\vert s_1\vert\,\vert s_2\vert}\textstyle\sum\limits_{i\in s_1}\textstyle\sum\limits_{j\in s_2}\ell(z_i, z_j;\theta)\, ,
\end{equation}
and we formally define DML as $\mathcal{A}_{[n]{\sxtimes}[n]}$, \ie, minimizing empirical expected loss with all possible pairs. We consider improving the generalization error of $\mathcal{A}_{s_1\sxtimes s_2}$ which is:
\begin{equation}\label{eq:gen_err}
\mathcal{L}(\mathcal{A}_{s_1\sxtimes s_2}) = \E{\ell(z_i,z_j; \mathcal{A}_{s_1\sxtimes s_2})}{z_i,z_j}.
\end{equation}

\section{Method}
\label{sec:method}
We will iteratively solve multiple proxy-based DML problems. At each problem, we re-initialize the class proxies by samples from the dataset. We relate the problems by regularizing the learned parameters to be in the close vicinity of the previous ones. In the following sections, we provide theoretical foundation behind the motivation of our method. We defer all the upcoming proofs to \refappendix.

We start with reformulating DML with a chance constraint. We will introduce two propositions that allow us to decompose the chance constraint into finite chance constraints. We also show minimizer of proxy-based DML satisfies some chance constraints. Hence, we link DML to finding a point in the intersection of finite sets, which we solve using iterative projections that correspond to regularized proxy-based DML problem instances.

\begin{figure}[!t]
\centering
  \centerline{\includegraphics[width=1.\linewidth,keepaspectratio]{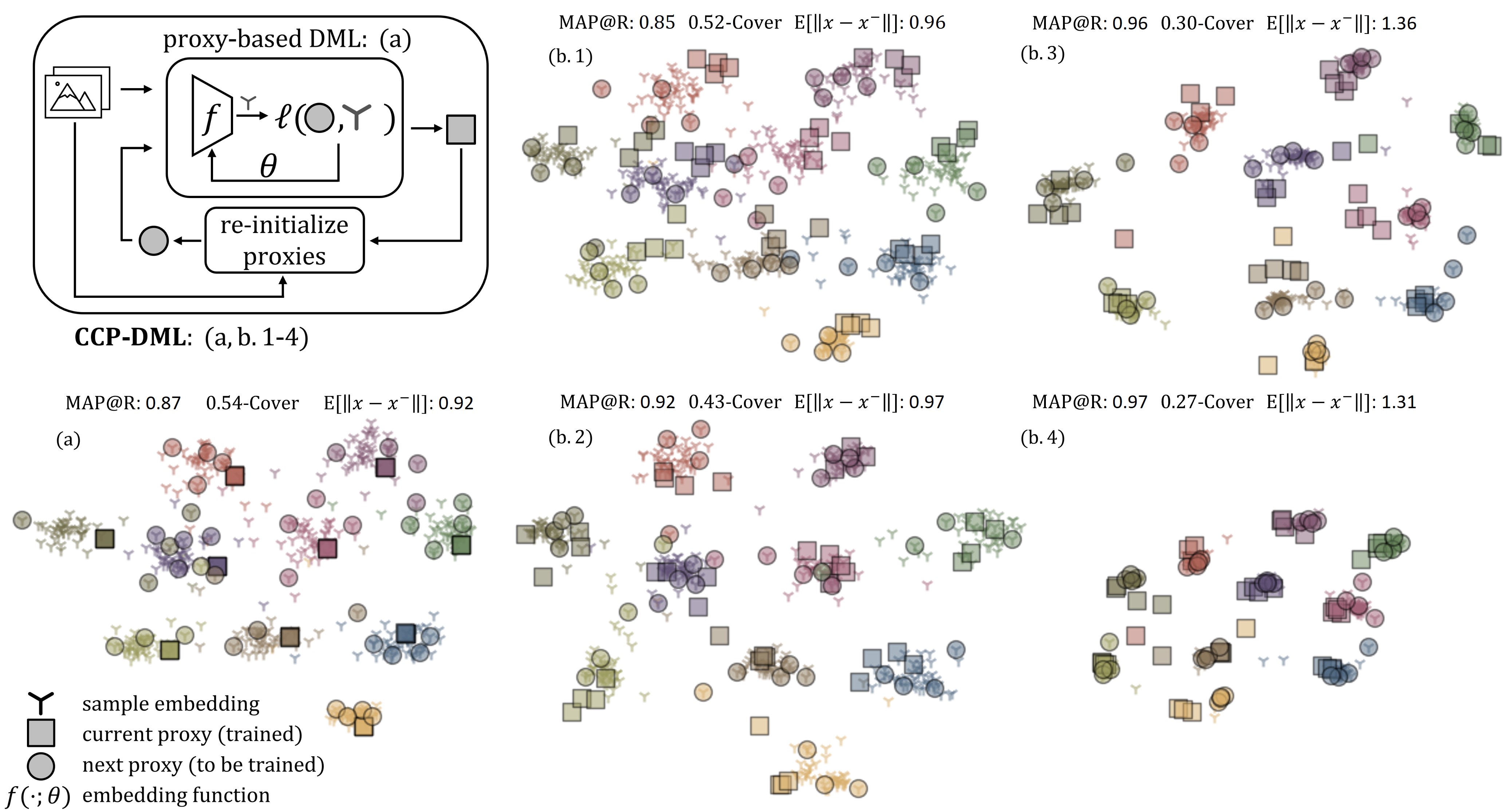}}
  \caption{Illustration of our method (CCP, \cref{algo:asap}) and the geometry of the embedding space in MNIST dataset: Boxes represent the converged proxies, while circles represent the next proxies resulting from $K$-Center (\cref{algo:kcenter}). (a) In proxy-based DML (before our method), proxies coalesce into one. (b) With CCP (through iterations 1-4), diverse proxies are obtained, resulting in a reduced covering radius. Magnified version is in suppl.\refmagnified}.
	\label{fig:radius}
\end{figure}
%
%
%
%
%
%
%

In the formulations throughout the paper, we rely on Lipschitz continuity of the loss function for which we refer to Lemma \ref{lemma}. Our approach focuses on enhancing the generalization performance in the seen domain, with implications for the crucial goal of generalizing to unseen classes in DML. In DML models, the embedding vector is derived by globally averaging local CNN features, which act as visual words \cite{zhou2016learning, gurbuz2019novel}. By prioritizing improved generalization during training, we can transfer the semantic knowledge captured by these visual words to effectively represent samples from  unseen classes. Furthermore, our empirical studies provide strong evidence supporting the effectiveness of the proposed formulations and their implications for DML.

\subsection{Chance Constrained Formulation of DML}
\label{sec:ccp_formulation}
We consider the solution of the following chance constrained feasibility problem:
\begin{equation}\label{eq:primary_ccp}
\min_\theta 0^\intercal\theta \,\,\, \text{s.\,to} \,\,\, p_{z_i,z_j}(y_{ij}(\Vert x_i \shortminus x_j \Vert_{f_\theta} - \beta)\geqslant 0) \leqslant \varepsilon
\end{equation}
with some small $\varepsilon$. In essence, we want the probability of observing two samples of the same (different) class being apart (close) more than $\beta$ in the embedding space being low. We write that probability as expected violation, $\E{\ind{y_{ij}(\Vert x_i \shortminus x_j \Vert_{f_\theta} - \beta)\geqslant 0}}{z_i,z_j}$ where $\ind{\cdot}$ being indicator function, and bound it for $ \beta \geqslant \alpha>0$ as:
\begin{equation} \label{eq:exp_bound}
\begin{split}
p_{z_i,z_j}(y_{ij}&(\Vert x_i \shortminus x_j \Vert_{f_\theta} - \beta)\geqslant 0) \\
 \leqslant
\nicefrac{1}{\alpha}&\E{(y_{ij}(\Vert x_i \shortminus x_j \Vert_{f_\theta} - \beta) + \alpha)_+}{z_i,z_j} \,,
\end{split}
\end{equation}
using Markov's inequality where $(u)_+=\max\lbrace0,u\rbrace$. Note that to each value of the expectation $e(\theta)$ there corresponds an $\varepsilon = \nicefrac{e(\theta)}{\alpha}$ which the chance constraint satisfies. Hence, we use the expectation as the surrogate of the penalty term for the chance constraint and can redefine the aforementioned feasibility problem as the expected loss minimization in \eqref{eq:DML_true} with $\ell(z_i,z_j;\theta) = (y_{ij}(\Vert x_i \shortminus x_j \Vert_{f_\theta} - \beta) + \alpha)_+$. In particular, we end up with minimization of the expected \textit{contrastive loss with positive margin} \cite{wu2017sampling}. 

We now consider the relaxed feasibility problem in which we consider $m$ chance constraints conditioned on given $m$ samples $\mathcal{S}{=}\lbrace z_i\rbrace_{i\in[m]}\sim p_\mathcal{Z}$, say \textit{anchor samples}. To be more precise, we want to find $\theta\in\mathcal{C}_\mathcal{S}$ with:
\begin{equation}\label{eq:constraint_sets}
\mathcal{C}_\mathcal{S}{=} \lbrace \theta \mid p_{z_j}(y_{ij}(\Vert x_i {\shortminus} x_j \Vert_{f_\theta} {-} \beta){\geqslant} 0) \leqslant \varepsilon, \, \forall i{\in}[m] \rbrace,
\end{equation}
where $[m]$ indexes the samples in $\mathcal{S}$. Using expectation bounds as in \eqref{eq:exp_bound}, the unconstrained problem becomes:
\begin{equation}\label{eq:relaxed_prob}
\theta^\ast = \argmin_{\theta} \tfrac{1}{m}\!\textstyle\sum\limits_{i\in[m]}\E{\ell(z_i,z_j;\theta)}{z_j}\, .
\end{equation}

We are particularly interested in the problem of the form in \eqref{eq:relaxed_prob} owing to its relation to proxy-based methods to characterize their generalization. Prior to delving into such a relation, we first bound the deviation from the actual expectation in \eqref{eq:DML_true} when we solve the problem in \eqref{eq:relaxed_prob} instead.
\begin{proposition}\label{p1}
Given $\mathcal{S}{=}\lbrace z_i\rbrace_{i\in[m]} {\overset{i.i.d.}{\sim}} p_\mathcal{Z}$ such that $\forall k {\in} \mathcal{Y}$ $\lbrace x_i{\mid} y_i{=}k\rbrace$ is $\delta_\mathcal{S}$-cover\footnote{$\mathcal{S}\subset \mathcal{S}^\prime$ is $\delta_\mathcal{S}$-cover of $\mathcal{S}^\prime$ if $\forall z^\prime {\in} \mathcal{S}^\prime$, $\exists z{\in} \mathcal{S} \,: \Vert z-z^\prime\Vert_2 \leqslant \delta_\mathcal{S}$.} of $\mathcal{X}$, $\ell(z_i,z_j;\theta)$ is \mbox{$\zeta$-Lipschitz} in $x_i$, $x_j$ for all $y_i$, $y_j$ and $\theta$, and bounded by $L$; then with probability at least $1-\gamma$,
\begin{equation*}
\begin{split}
\Big\vert\E{\ell(z_i,z_j;\theta)}{z_i,z_j} &- \tfrac{1}{m}\!\textstyle\sum\limits_{i\in[m]}\E{\ell(z_i,z_j;\theta)}{z_j}\Big\vert \\
&\leqslant \mathcal{O}(\zeta\,\delta_\mathcal{S}) + \mathcal{O}(L\,\sqrt{\nicefrac{\log\tfrac{1}{\gamma}}{m}}).
\end{split}
\end{equation*}
\end{proposition}
Proposition \ref{p1} gives an upper bound which is controlled by the diversity of the anchor samples defining the relaxed problem. Theoretically, such a controlled bound allows DML to be formulated as a feasibility problem of finite sets for some accepted error tolerance. In practice, the best we can do is using all the samples in the dataset as the anchor samples when defining $\mathcal{C}_\mathcal{S}$ in \eqref{eq:constraint_sets}. Granted that the minimization in \eqref{eq:relaxed_prob} with the empirical loss boils down to the classical DML in \eqref{eq:empi}, it has different stochastic optimization procedure. The relaxed problem suggests sampling batch of instances rather than pairs, which yields less noisy gradient estimates with the same batch budget.

\subsection{Reducing Chance Constraints}
The loss terms conditioned on anchor samples in \eqref{eq:relaxed_prob} are computationally prohibitive in large-scale problems. Thus, we are interested in reducing the chance constraints, \ie, anchor samples. To this end, proxy-based methods are quite related in that a proxy-based DML constitutes a superset of the feasible region of the primary DML problem in \eqref{eq:primary_ccp} as we will show shortly.

Proxy-based methods use parametric vectors $\lbrace\rho_i\rbrace_{i\in[C]}$ to represent embedding of the class centers and minimize the pair losses with respect to those centers. Formally, given a dataset $\lbrace z_i \rbrace_{i\in[n]}\sim p_\mathcal{Z}$, proxy-based methods consider the following problem:
\begin{equation}\label{eq:proxy_dml}
\min_{\theta,\rho} \tfrac{1}{n\,C}\textstyle\sum\limits_{i\in[C]}\sum\limits_{j\in[n]}\hat{\ell}(\rho_i, z_j; \theta),
\end{equation}
where $\hat{\ell}(\rho_i, z_j; \theta)$ is a loss term in which the pairwise distance is computed as $\Vert \rho_i-f(x_j;\theta)\Vert_2$. We can associate an algorithm $\mathcal{A}_{s\sxtimes[n]}$ defined in \eqref{eq:empi} to the minimizer of \eqref{eq:proxy_dml} with $e(\mathcal{A}_{s\sxtimes[n]})$ training error where $s=\lbrace s(i)\in [n]\mid f(x_{s(i)};\mathcal{A}_{s\sxtimes[n]})=\rho_i\rbrace_{i\in[C]}$. In other words, to each proxy, we associate a dataset sample whose embedding matches that proxy, assuming such sample exists. Hence, the minimizer of the proxy-based methods can be reformulated as the following feasibility problem: 
\begin{equation}\label{eq:proxy_constraint}
\!\!\!\!\min_\theta 0^\intercal \theta \,\,  \text{s.\,to} \,\,  p_{z_j}(y_{ij}(\Vert x_{i}{\shortminus} x_j\Vert_{f_\theta} {-} \beta){\geqslant} 0) \leqslant \varepsilon, \, \forall i{\in}s,
\end{equation}
where $s$, as explained above, indexes $C$-many dataset samples corresponding to proxies, and $\varepsilon {=} \tfrac{1}{\alpha}\mathcal{L}(\mathcal{A}_{s\sxtimes[n]})$ from the expression in \eqref{eq:exp_bound}. $\mathcal{L}(\mathcal{A}_{s\sxtimes[n]})$ defined in \eqref{eq:gen_err} is shown to be bounded in \cite{bellet2015robustnessML}, hence so is $\varepsilon$. Reformulation of proxy-based DML defines the feasibility problem in \eqref{eq:constraint_sets} with one sample per class.

We now consider more general case where we use $m$ samples per class from the dataset $\lbrace z_i \rbrace_{i\in[n]}{\sim} p_\mathcal{Z}$ to define the feasibility problem. We have $m$-many disjoint 1-per-class sets $s=\cup_{k\in[m]} s_k$, where $s_k=\lbrace s_k(i){\in} [n] \mid y_{s_k(i)}{=}i  \rbrace_{i\in[C]}$ with $\cap_{k\in[m]} s_k=\emptyset$. We define the problem:
\begin{equation}\label{eq:reduced_feasibility}
\begin{split}
&\min_{\theta\in\cap_k\mathcal{C}_{s_k}}0^\intercal\theta\quad\text{where} \quad \mathcal{C}_{s_k} = \{ \theta \mid \forall i\in s_k, \\
&p_{z_j}(y_{ij}\Vert x_i\shortminus x_j\Vert_{f_\theta} - \beta)\geqslant 0) \leqslant \varepsilon \}.
\end{split}
\end{equation}
Solving the problem by minimizing the empirical expectation bounds in \eqref{eq:relaxed_prob}, we end up with an algorithm $\mathcal{A}_{s\sxtimes[n]}$ in which we are minimizing expected loss over a subset of all possible pairs. We want to characterize the generalization performance of the algorithm $\mathcal{A}_{s\sxtimes[n]}$. We consider the following bound from \cite{sener2018active} for the generalization error:
\begin{equation}\label{eq:bounding_error}
\begin{split}
&\E{\ell(z_i,z_j;\mathcal{A}_{s\sxtimes[n]})}{z_i,z_j}
\leqslant \Big\vert\tfrac{1}{\vert s \vert\,n}\!\!\!\!\!\!\!\textstyle\sum\limits_{i,j\in s\sxtimes[n]}\!\!\!\!\!\! \ell(z_i,z_j;\mathcal{A}_{s\sxtimes[n]})\Big\vert_{(\mathcal{L}_1)} \\
&+\Big\vert \E{\ell(z_i,z_j;\mathcal{A}_{s\sxtimes[n]})}{z_i,z_j} \!-\! \tfrac{1}{n^2}\!\!\!\!\!\!\!\!\textstyle\sum\limits_{i,j\in[n]\sxtimes[n]}\!\!\!\!\!\!\! \ell(z_i,z_j;\mathcal{A}_{s\sxtimes[n]}) \Big\vert_{(\mathcal{L}_2)} \\
&+\Big\vert \tfrac{1}{n^2}\!\!\!\!\!\!\!\textstyle\sum\limits_{i,j\in[n]\sxtimes[n]} \!\!\!\!\!\!\! \ell(z_i,z_j;\mathcal{A}_{s\sxtimes[n]}) - 
\tfrac{1}{\vert s \vert\,n}\!\!\!\!\!\!\!\textstyle\sum\limits_{i,j\in s\sxtimes[n]}\!\!\!\!\!\! \ell(z_i,z_j;\mathcal{A}_{s\sxtimes[n]}) \Big\vert_{(\mathcal{L}_3)} 
\end{split}
\end{equation}
where the bound is controlled by ($\mathcal{L}_1$) training loss (\ie, $e(\mathcal{A}_{s\sxtimes[n]})$), ($\mathcal{L}_2$) the deviation between expected loss and empirical loss over all possible pairs, and ($\mathcal{L}_3$) the deviation between empirical loss over all possible pairs and empirical loss over the subset of pairs defining the algorithm $\mathcal{A}_{s\sxtimes[n]}$. It is widely observed that high capacity CNNs can reach very small training error. Moreover, $\mathcal{L}_2$ is proved to be bounded in \cite{bellet2015robustnessML} and is independent of $\mathcal{A}$. Thus, $\mathcal{L}_3$ characterizes the generalization performance of using the subset of pairs over exploiting all possible pairs. 
\begin{proposition} \label{p2}
Given $\lbrace z_i\rbrace_{i\in[n]} \overset{i.i.d.}{\sim} p_\mathcal{Z}$ and a set $s\subset[n]$. If $s=\cup_k s_k^\prime$ with $s_k^\prime$ is the $\delta_s$-cover of $\lbrace i\in [n] \mid y_i=k\rbrace$ (\ie, the samples in class $k$ ), $\ell(z_i,z_j;\theta)$ is $\zeta$-Lipschitz in $x_i, x_j$ for all $y_i, y_j$ and $\theta$, and bounded by $L$,  $e(\mathcal{A}_{s\sxtimes[n]})$ training error; then with probability at least $1-\gamma$ we have: 
\begin{equation*}
\begin{split}
\Big\vert \tfrac{1}{n^2}\!\!\!\!\!\!\!\textstyle\sum\limits_{i,j\in[n]\sxtimes[n]} \!\!\!\!\!\!\! &\ell(z_i,z_j;\mathcal{A}_{s\sxtimes[n]}) - 
\tfrac{1}{\vert s \vert\,n}\!\!\!\!\!\!\!\textstyle\sum\limits_{i,j\in s\sxtimes[n]}\!\!\!\!\!\! \ell(z_i,z_j;\mathcal{A}_{s\sxtimes[n]}) \Big\vert \\
&\leqslant
\mathcal{O}(\zeta\,\delta_s) + \mathcal{O}(e(\mathcal{A}_{s\sxtimes[n]})) + \mathcal{O}(L\,\sqrt{\tfrac{\log\nicefrac{1}{\gamma}}{n}}).
\end{split}
\end{equation*}

\end{proposition}
\begin{corollary}\label{corr}
Generalization of the proxy-based methods can be limited by the maximum of distances between the proxies and the corresponding class samples in the dataset.
\end{corollary}

Proposition \ref{p2} implies that increasing the number of chance constraints with more anchor samples in the feasible point problem formulation of DML improves the generalization error bound as long as the included samples improve the covering radius of the dataset. In other words, including more anchor samples do not improve the bound unless the covering radius is decreased. Similarly, Corollary \ref{corr} informally suggests possible improvement on the generalization error bound of the proxy-based methods if we manage to introduce more proxies which are spread over the dataset once trained. In practice introducing more proxies generally does not help the performance since they eventually coalesce into a single point \cite{Qian_2019_ICCV}. Besides, the computation resource limits the number of proxies to be included in the formulation. In the next section, we develop an approach to alleviate these problems.

\subsection{Solving the Feasibility Problem}
We now introduce our chance constrained programming (CCP) method, outlined in Algorithm \ref{algo:asap}, exploiting proxy-based training together with satisfying arbitrarily increased chance constraints. In short, we repeatedly solve a proxy-DML and improve the solution by re-initializing the proxies with the new samples reducing the covering radius.

We consider the problem in \eqref{eq:reduced_feasibility} as finding a point in the intersection of the sets. In particular, given dataset $\lbrace z_i \rbrace_{i\in[n]}{\sim} p_\mathcal{Z}$, we have $m$ many 1-per-class sets \mbox{$s_k=\lbrace s_k(i)\in [n] \mid y_{s_k(i)}=i  \rbrace_{i\in[C]}$} to define the constraint set as $\mathcal{C}_s = \displaystyle\cap_{k\in[m]} \mathcal{C}_{s_k}$. If the sets were closed and convex, the problem would be solvable by iterative projection methods \cite{bregman1967relaxation, bauschke2000dykstras}. Nevertheless, it is not uncommon to perform iterative projection methods to non-convex set intersection problems \cite{pang2015nonconvex, solomon2015convolutional}. Hence, we propose to solve the problem approximately by performing iterative projections onto the feasible sets $\mathcal{C}_{s_k}$ defined by $s_k$. At each iteration $k$ we solve the following projection problem given $\theta^{(k\shortminus 1)}$:

\renewcommand{\algorithmiccomment}[1]{\hfill\eqparbox[h]{COMMENT}{// #1}}
\renewcommand{\algorithmicrequire}{initialize}
\renewcommand{\algorithmicensure}{\textbf{repeat}}

\begin{algorithm}[t]
\caption{CCP DML}\label{algo:asap}
\begin{algorithmic}
\Require $\theta^{\ast}$ randomly, given $\lbrace z_i\rbrace_{i\in[n]}{\sim}p_\mathcal{Z}$ dataset
\Require $\rho^\ast$ with random samples, set budget $b$
\Ensure 
	\State $\rho\gets GreedyKCenterProxy(\rho^\ast,b, f(\cdot; \theta^{\ast}))$
	\Repeat
		\State sample $s=\lbrace j\in[n] \rbrace_{j\in[m]}{\sim}[n]$ a batch
		\State $g_\theta{\gets}\lambda\,(\theta^{\ast} {-} \theta ) {+} \nabla_{\theta}\tfrac{1}{m\vert\rho\vert}\sum_{\rho\sxtimes s}\ell(\rho_i, z_j; \theta)$
		\State $g_\rho{\gets} \nabla_{\rho}\tfrac{1}{m\vert\rho\vert}\sum_{\rho\sxtimes s}\ell(\rho_i, z_j; \theta)$
		\State $(\theta, \rho){\gets}ApplyGradient(\theta, \rho, g_\theta, g_\rho)$
\Until {convergence}
\State $\theta^{\ast}{\gets}\theta$, $\rho^\ast{\gets}\rho$
\end{algorithmic}
\textbf{until} convergence
\end{algorithm}
\begin{equation}\label{eq:alt_proj}
\theta^{(k)}=\argmin_{\theta\in\mathcal{C}_{s_k}} \tfrac{1}{2}\Vert\theta^{(k\shortminus 1)}-\theta\Vert_2^2,
\end{equation}
where ${C}_{s_k}$ is defined in \eqref{eq:reduced_feasibility}. Using expectation bounds as the surrogate of the penalty terms for the chance constraints as we do in \cref{sec:ccp_formulation}, we have: 
\begin{equation}\label{eq:alt_proj_unconstrained}
\theta^{(k)}\!=\!\argmin_{\theta} \tfrac{\lambda}{2}\Vert\theta^{(k\shortminus 1)}\shortminus\theta\Vert_2^2 + \tfrac{1}{C}\!\!
 \textstyle\sum\limits_{i\in [C]}\!\!\E{\ell(z_{s_k(i)},z_j;\theta)}{z_j},
\end{equation}
where $\lambda$ is a hyperparameter for the projection regularization. We can minimize the resultant loss by using batch stochastic gradient approaches. However, the batch should be augmented by $C$ many anchor samples to compute the loss, which becomes prohibitive for large-scale problems. To alleviate costly embedding computation of $C$ many samples, we propose to use proxies $\rho_i$ in place of the embedding of the samples $z_{s_k(i)}$. Namely, at each iteration $k$, we initialize $\rho_i = f(z_{s_k(i)};\theta^{(k\shortminus 1)})$ and solve:
\begin{equation}\label{eq:alt_proj_unconstrained_proxy}
\theta^{(k)}\!, \rho^{\ast}\!\!=\!\argmin_{\theta, \rho} \tfrac{\lambda}{2}\Vert\theta^{(k\shortminus 1)}\shortminus\theta\Vert_2^2 + \tfrac{1}{C}\!\!
 \textstyle\sum\limits_{i\in [C]}\!\!\E{\ell(\rho_i,z_j;\theta)}{z_j},
\end{equation}
where the resultant problem we solve at each iteration corresponds to a proxy-based DML. Any pairwise distance based loss can replace $\ell(\cdot)$ with anchor samples being class proxies. \Ie, we repurpose existing objectives with a regularization term in an iterative manner. Although we set up the formulation using single proxy per class, extending it to accommodate multiple proxies is a straightforward process.

Theoretically, we should cycle through the sets until convergence to solve $\theta{\in}\displaystyle\cap_{k\in[m]} \mathcal{C}_{s_k}$. Thus, we must pick anchor samples for each set to initialize proxies. The updates of the proxies are not guaranteed to mimic the actual updates of the corresponding anchor samples. With that being said, we will still have a solution, as \eqref{eq:proxy_constraint} suggests, to feasibility of some chance constraints as long as the converged proxies $\rho^{\ast}$ are diverse. We empirically observe that the proxies initialized with diverse samples converge to embedding of distinct samples (\cref{fig:radius}). Hence, on one hand, we have solutions to different constraint sets as long as we re-initialize the proxies with new samples and solve proxy-based DML problems. On the other hand, Proposition \ref{p2} implies that generalization is improved as long as we end up with converged proxies reducing the covering radius. Therefore, the theory suggests a set intersection mechanism to reduce the covering radius yet allows a greedy algorithm via iterative projections to select (\ie, initialize) the next proxies on the fly instead of explicitly defining the sets we will iterate on. Such a result is useful especially for the cases where the dataset is stochastically extended with random data augmentations which obstruct explicit set forming.

\textbf{Proxy selection.} We can simply use random sampling for anchor samples to initialize proxies since we eventually observe informative samples reducing the covering radius through the iterations. We can as well explicitly mine samples that possibly help with reducing the covering radius. Thus, we also exploit clever selection of proxies as outlined in Algorithm \ref{algo:kcenter}. Given a budget $b$, we sample $b$ many instances per class and compute their embeddings to form a pool. We then select the samples that reduce the covering radius most once added to proxy set. This selection is equivalent to $K$-Center problem as formulated in \cite{sener2018active}. Such a selection of proxies helps converged proxies to be diverse. $b=1$ reduces to random sampling. In both, we inherently increase the number of anchor samples defining the problem and hence reducing the covering radius.


\renewcommand{\algorithmiccomment}[1]{\hfill\eqparbox[h]{COMMENT}{// #1}}
\renewcommand{\algorithmicrequire}{\textbf{input:}}
\renewcommand{\algorithmicensure}{\textbf{repeat}}



\begin{algorithm}[H]
\caption{Greedy $K$-Center Proxy}\label{algo:kcenter}
\begin{algorithmic}
\Require proxy set $\rho$, sampling budget $b$ and $f(\cdot;\theta)$
\Ensure for each class $c$
	\State $s_c\gets\lbrace x_i \mid y_i=c\rbrace_{i\in[b]}$, $b$-sample-per-class
	\State initialize $r_c\gets\lbrace\rbrace$, $p\gets f(s_c;\theta)$
	\Repeat
		\State $q\gets\argmax_{u\in p \setminus r_c}\min_{v\in\rho_c\cup r_c}\Vert u-v \Vert_2$
		\State $r_c \gets \lbrace q\rbrace \cup r_c$
	\Until {$\vert r_c\vert=\vert \rho_c\vert$}
\end{algorithmic}
\textbf{return} $\cup_c r_c$
\end{algorithm}



\subsection{Implementation Details}
\label{sec:implementation}
\textbf{Embedding function.} For the embedding function $f(\cdot;\theta)$ we use ImageNet \cite{russakovsky2015imagenet} pretrained CNNs with ReLU activation, max- and average-pooling. We exploit architectures until the output of the global average pooling layer. We add a fully connected layer (\ie, linear transform) to the output of the global average pooling layer to obtain the embedding vectors. We state the following lemma to prove our loss is Lipschitz continuous.
\begin{lemma}\label{lemma}
Generalized contrastive loss defined as $\ell(z_i,z_j;\theta)\coloneqq(y_{ij}(\Vert x_i\shortminus x_j\Vert_{f_\theta} - \beta) + \alpha)_+$ is $\sqrt{2}\omega^L$-Lipschitz in $x_i$ and $x_j$ for all $y_i,y_j,\theta$ for the embedding function $f(\cdot;\theta)$ being $L$-layer CNN (with ReLU, max-pool, average-pool) with a fully connected layer at the end, where $\omega$ is the maximum sum of the input weights per neuron.
\end{lemma}
$\omega$ can be made arbitrarily small with weight regularization, which is commonly used \cite{wang2020cross}. SOTA methods widely use $\ell2$ normalization on the embeddings. For normalization, we apply $\hat{v} = \nicefrac{v}{\Vert v \Vert_2}$ if $\Vert v \Vert_2 \geqslant 1$ or identity otherwise (\ie, $\hat{v} = v$ if $\Vert v \Vert_2 \leqslant 1$). Unlike $\ell2$ normalization, such a transform is Lipschitz continuous, hence so are our loss. 

\textbf{Solving projections.} Performing a projection defined in \eqref{eq:alt_proj_unconstrained_proxy} involves a minimization problem. We monitor MAP@R validation accuracy and use early stopping patience of 3 to pass the next projection.
\section{Experimental Work}

\subsection{Setup}
\label{sec:setup}
We follow the suggestions of recent work \cite{roth2020revisiting, musgrave2020metric, fehervari2019unbiased} explicitly studying the fair evaluation strategies for DML  in order to minimize the confounding of the factors other than our method. Specifically, we mostly follow the MLRC procedures proposed in \cite{musgrave2020metric} to provide fair and unbiased evaluation of our method as well as comparisons with the other methods. We offer detailed experimental setup information in the supplementary material \refexpdetail for reproducibility.
\\
\textbf{Backbone.} BN-Inception \cite{normalization2015accelerating}  with 128D embedding.
\\
\textbf{Datasets.} CUB \cite{wah2011caltech}, Cars \cite{krause2014submodular}, In-shop \cite{liu2016deepfashion}, SOP \cite{oh2016deep} with MLRC \cite{musgrave2020metric} data augmentation.
\\
\textbf{Training.} Adam \cite{kingma2014adam} optimizer with $10^{\shortminus 5}$ learning rate, $10^{\shortminus 4}$ weight decay, 32 batch size (4  per class), 4-fold: 4 models (1 for each $\nicefrac{3}{4}$ train set partition).
\\
\textbf{Evaluation.} Average performance (Separated-128D) with mean average precision (MAP@R) at R where R is defined for each query and is the total number of its true references.
\\
\textbf{Losses with CCP.} \textit{C1-CCP}: Contrastive \cite{hadsell2006dimensionality}, \textit{C2-CCP}: Contrastive with positive margin \cite{wu2017sampling}, \textit{MS-CCP}: Multi-similarity (MS) \cite{Wang_2019_CVPR_MS}, \textit{Triplet-CCP}: Triplet \cite{schroff2015facenet}. 
\\
\textbf{Compared methods and fairness.} We compare our method against proxy-based SoftTriple \cite{Qian_2019_ICCV}, ProxyAnchor \cite{kim2020proxy} and ProxyNCA++ \cite{teh2020proxynca++} methods as well as XBM \cite{wang2020cross}. Our experiments cover wide range of the DML losses since ProxyAnchor is indeed proxy-based MS loss except for missing a margin term, similarly ProxyNCA is $\log\Sigma\exp$-approximation of proxy-based Triplet with hard negative mining, and for single proxy case SoftTriple $\equiv$ ProxyNCA. 
\\
\textbf{CCP hyperparameters.} We introduce
3 new hyperparameters to a typical DML: $\lambda$, \#$proxy$ (proxy per class),  $b$ (pool size). We optimize $\lambda$-\#$proxy$ with Bayesian search (details are in supplementary material \refablation) and $b$-\#$proxy$ with grid search (\cref{fig:pool_size}). Based on our analysis, we set $\lambda{=}2\cdot10^{\shortminus 4}$, \#$proxy{=}8$, $b{=}12$ for CUB and Cars. For SOP and In-shop, we reduce \#$proxy{=}4$ and $b{=}7$ owing to relatively less number of samples per class in the dataset.
\\
\textbf{Conventional evaluation.} We additionally follow the relatively old-fashioned \textit{conventional} procedure \cite{oh2016deep} for the evaluation of our method. We use BN-Inception \cite{normalization2015accelerating} and ResNet50 \cite{he2016identity}  backbones with 512D embeddings. We use global max pooling as well as global average pooling, likewise the recent approaches \cite{venkataramanan2022it,teh2020proxynca++,kim2020proxy, wang2020cross}. We use batch size of 128 for BN-Inception and 96 
for ResNet50.



\subsection{Results}

\begin{figure}[b]
\centering
\centerline{\includegraphics[width=1.0\linewidth,keepaspectratio]{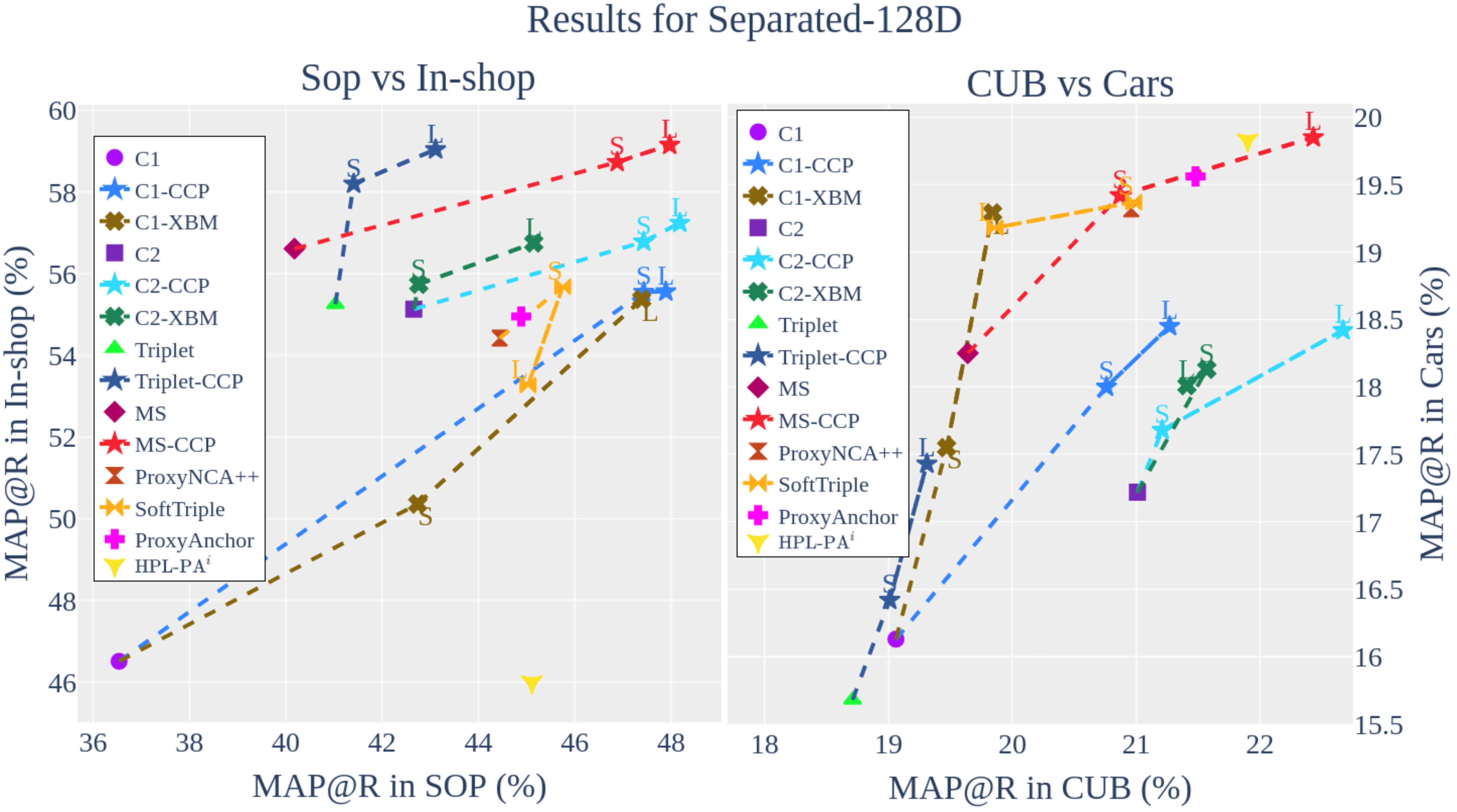}}




\caption{Summary of relative improvements for MLRC evaluation. \textsuperscript{i}\textit{In-shop result is not available for HPL-PA \cite{yang2022hierarchical}}.}
\label{fig:comparisons}
\end{figure}
\begin{table}[!t]
\centering
\caption{Conventional evaluation with BN-Inception. Red: the best. Blue: the second best. Bold: previous SOTA.}
\label{tab:conventional_bninception}
\resizebox{\linewidth}{!}{%
\begin{tabular}{@{}ccccc@{}}
\toprule
\multicolumn{1}{l}{\textbf{Backbone $\rightarrow$}}     & \multicolumn{4}{c}{\textbf{BN-Inception-512D}}                                                                                                                \\ \midrule
\multicolumn{1}{l}{\textbf{Dataset $\rightarrow$}}      & \textbf{CUB}                          & \textbf{Cars196}                      & \textbf{SOP}                          & \textbf{In-shop}                      \\ \cmidrule(l){2-5} 
\multicolumn{1}{l}{\textbf{Method $\downarrow$}}        & \textbf{R@1}                          & \textbf{R@1}                          & \textbf{R@1}                          & \textbf{R@1}                          \\ \midrule
\multicolumn{1}{l|}{SoftTriple-L \cite{Qian_2019_ICCV}} & 65.40                                 & 84.50                                 & 78.60                                 & -                                     \\
\multicolumn{1}{l|}{C1-XBM-L \cite{wang2020cross}}      & 65.80                                 & 82.00                                 & 79.50                        & 89.90                                 \\
\multicolumn{1}{l|}{ProxyAnchor \cite{kim2020proxy}}    & 68.40                                 & 86.10 & 79.10                                 & \textbf{91.50}                        \\
\multicolumn{1}{l|}{DiVA \cite{milbich2020diva}}        & 66.80                                 & 84.10                                 & 78.10                                 & -                                     \\
\multicolumn{1}{l|}{ProxyFewer \cite{zhu2020fewer}}     & 66.60                                 & 85.50                                 & 78.00                                 & -                                     \\
\multicolumn{1}{l|}{PROFS \cite{profs}}    & 66.00                        & 86.30 & 78.70                                 & -
\\
\multicolumn{1}{l|}{Margin-S2SD \cite{roth2021s2sd}}    & \textbf{68.50}                        & {\color[HTML]{3166FF} \textbf{87.30}} & 79.30                                 & -                                     \\ 
\multicolumn{1}{l|}{HIST \cite{lim2022hypergraph}}    & {\color[HTML]{3166FF}\textbf{69.70}}                        & {\color[HTML]{FE0000} \textbf{87.40}} & \textbf{79.60}                                 & -                                     \\\midrule
\multicolumn{1}{l|}{\textbf{C1-CCP-L}}                           & 67.74                                 & 83.74                                 & {\color[HTML]{3166FF} \textbf{79.86}} & 90.98                                 \\
\multicolumn{1}{l|}{\textbf{C2-CCP-L}}                           & {\color[HTML]{FE0000} \textbf{69.87}} & 83.90                                 & {\color[HTML]{FE0000} \textbf{80.01}} & {\color[HTML]{3166FF} \textbf{91.72}} \\
\multicolumn{1}{l|}{\textbf{MS-CCP-L}}                                                & 69.09 & 86.01                                 & 79.75                                 & {\color[HTML]{FE0000} \textbf{91.84}} \\ \bottomrule
\end{tabular}%
}
\end{table}

\begin{table}[!ht]
\centering
\caption{Conventional evaluation with ResNet50. Red: the best. Blue: the second best. Bold: previous SOTA. {\footnotesize\textsuperscript{$\dagger$}\textit{Results from LIBC} \cite{intrabatch}.}}
\label{tab:conventional_resnet}
\resizebox{\linewidth}{!}{%
\begin{tabular}{@{}ccccc@{}}
\toprule
\multicolumn{1}{l}{\textbf{Backbone $\rightarrow$}}                    & \multicolumn{4}{c}{\textbf{ResNet50}}                                                                                                                         \\ \midrule
\multicolumn{1}{l}{\textbf{Dataset $\rightarrow$}}                     & \textbf{CUB}                          & \textbf{Cars196}                      & \textbf{SOP}                          & \textbf{In-shop}                      \\ \cmidrule(l){2-5} 
\multicolumn{1}{l}{\textbf{Method $\downarrow$}}                       & \textbf{R@1}                          & \textbf{R@1}                          & \textbf{R@1}                          & \textbf{R@1}                          \\ \midrule
\multicolumn{1}{l|}{C1-XBM$^{128}$   \cite{wang2020cross}}             & -                                     & -                                     & 80.60                                 & 91.30                                 \\
\multicolumn{1}{l|}{ProxyAnchor$^{512}$ \cite{kim2020proxy}}           & 69.70                                 & 87.70                                 & 80.00\textsuperscript{$\dagger$} & 92.10\textsuperscript{$\dagger$} \\
\multicolumn{1}{l|}{DiVA$^{512}$ \cite{milbich2020diva}}               & 69.20                                 & {\color[HTML]{000000} 87.60}          & 79.60                                 & \textbf{-}                            \\
\multicolumn{1}{l|}{ProxyNCA++$^{512}$   \cite{teh2020proxynca++}}     & 66.30                                 & 85.40                                 & 80.20                                 & 88.60                                 \\
\multicolumn{1}{l|}{PROFS$^{512}$ \cite{profs}}    & 64.90                        & 81.30 & 76.90                                 & -
\\
\multicolumn{1}{l|}{Margin-S2SD$^{512}$ \cite{roth2021s2sd}}           & 69.00                                 & {\color[HTML]{3166FF} \textbf{89.50}} & 81.20                                 & -                                     \\
\multicolumn{1}{l|}{LIBC$^{512}$ \cite{intrabatch}}                    & \textbf{70.30}                        & 88.10                                 & \textbf{81.40}                        & {\color[HTML]{FE0000} \textbf{92.80}} \\
\multicolumn{1}{l|}{ProxyAnchor-DIML$^{128}$   \cite{zhao2021towards}} & 66.46                                 & {\color[HTML]{000000} 86.13}          & 79.22                                 & -                                     \\

\multicolumn{1}{l|}{MS+Metrix$^{512}$ \cite{venkataramanan2022it}} & {\color[HTML]{FE0000} \textbf{71.40}} & {\color[HTML]{FE0000} \textbf{89.60}} & 81.00                                         &  92.20 \\
\multicolumn{1}{l|}{HIST$^{512}$ \cite{lim2022hypergraph}}    & {\color[HTML]{FE0000}\textbf{71.40}}                        & {\color[HTML]{FE0000} \textbf{89.60}} & \textbf{81.40}                                 & -       
\\
\multicolumn{1}{l|}{HPL-PA$^{512}$ \cite{yang2022hierarchical}}    & -                        & - & 80.04                                 & 92.46       
\\
\midrule
\multicolumn{1}{l|}{\textbf{C1-CCP-L}$^{512}$}                                  & 69.87                                 & 87.12                                 & {\color[HTML]{FE0000} \textbf{81.74}} & 92.07                                 \\
\multicolumn{1}{l|}{\textbf{C2-CCP-L}$^{512}$}                                  & {\color[HTML]{3166FF} \textbf{71.04}} & 85.93                                 & {\color[HTML]{3166FF} \textbf{81.66}} & {\color[HTML]{000000} 92.46}          \\
\multicolumn{1}{l|}{\textbf{MS-CCP-L}$^{512}$}                                  & {70.37} & {89.02} & 81.59                                 & {\color[HTML]{3166FF} \textbf{92.71}} \\ \bottomrule
\end{tabular}%
}
\end{table}

\textbf{MLRC.} We present the tabulated MLRC evaluation results in \refmlrctab and summarize MAP@R rankings with 128D embeddings in \cref{fig:comparisons}. We use Method-S/L naming convention to denote memory size in XBM, and the proxy per class in SoftTriple and CCP where S denotes 1, and L denotes 4(10) for SoftTriple and 4(8) for CCP in In-shop, SOP (CUB, Cars196). For fairness, we match XBM memory size and the number of proxies in CCP. We observe that CCP consistently outperforms the associated baseline methods on each dataset. Contrastive loss' compelling performance with CCP is important to support the implications of our formulation. Moreover, performance improvements on the losses which do not directly fit in our formulation show the broader applicability of our method to the pairwise distance based losses. Additionally,  CCP framework outperforms not only the related proxy-based methods but also every single benchmarked approach in \cite{musgrave2020metric}. When compared with SoftTriple and XBM (\ie, multiple proxy methods), CCP outperforms them by large margin especially in the cases where less number of proxies are used (\ie, \textit{method}-S comparisons in \cref{fig:comparisons}). We observe especially in large-scale datasets (SOP \& In-shop) that even single proxy per class brings substantial performance improvement with CCP. Finally, outperforming hierarchical proxy-based loss \cite{yang2022hierarchical} further supports CCP's superior embedding geometry. 

\textbf{Conventional.} We provide R@1 results in \Cref{tab:conventional_bninception,tab:conventional_resnet} for the comparison with SOTA. We observe that our method outperforms SOTA in most cases and performs on par with or slightly worse in a few. Predominantly, our method has superior performance on large-scale datasets, especially compared to  PROFS \cite{profs} which suffers from the poor scalability of exploiting class representatives. 



\subsection{Ablations}
\label{sec:ablation}
We include the analyses for the implications of our formulation and the effects of the hyperparameters. We defer computational analysis to supplementary material \refablation.

\textbf{Proof of the concept.} We evaluate our method using ResNet20V2 \cite{he2016identity} on MNIST \cite{mnist} dataset with 2D embeddings to show the implications of our formulation. In \cref{fig:radius}, we provide the distribution of the samples in the embedding space. We use 4 proxies per class and pool size $b{=}16$. We observe that when single proxy-based method is converged (\cref{fig:radius}-(a)), the class proxies collapse to a single point. Once we continue training with proposed approach (\cref{fig:radius}-(b)), the covering radius decreases, leading to performance improvement. We as well observe that diverse samples result in diverse proxies. In supplementary material \refablation, we extend this study and further provide the visualization of the validation data in CUB dataset to see how reducing the covering radius in training transfers to the test domain. We additionally experiment the case where we use samples instead of proxies. Though it is not practically applicable to large-scale problems, it is important to see whether our intuitions about alternating proxies in place of samples hold. We obtain $98.06\%$  MAP@R performance with sample-based training against $97.21\%$ MAP@R performance of proxy-based training. This empirical result supports our motivation on using the proxies in place of samples.


\begin{figure}[t]
\centering
\begin{minipage}{.49\linewidth}

  \centerline{\includegraphics[width=1.0\linewidth,keepaspectratio]{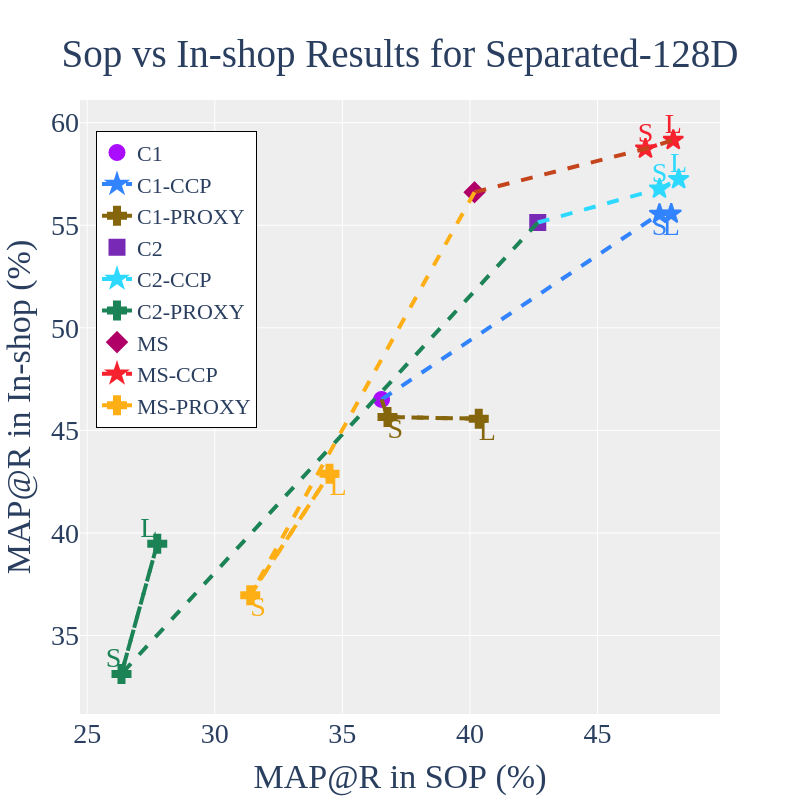}}
  \centerline{\scriptsize{(a)}}

\end{minipage}
\begin{minipage}{.49\linewidth}

  \centerline{\includegraphics[width=1.0\linewidth,keepaspectratio]{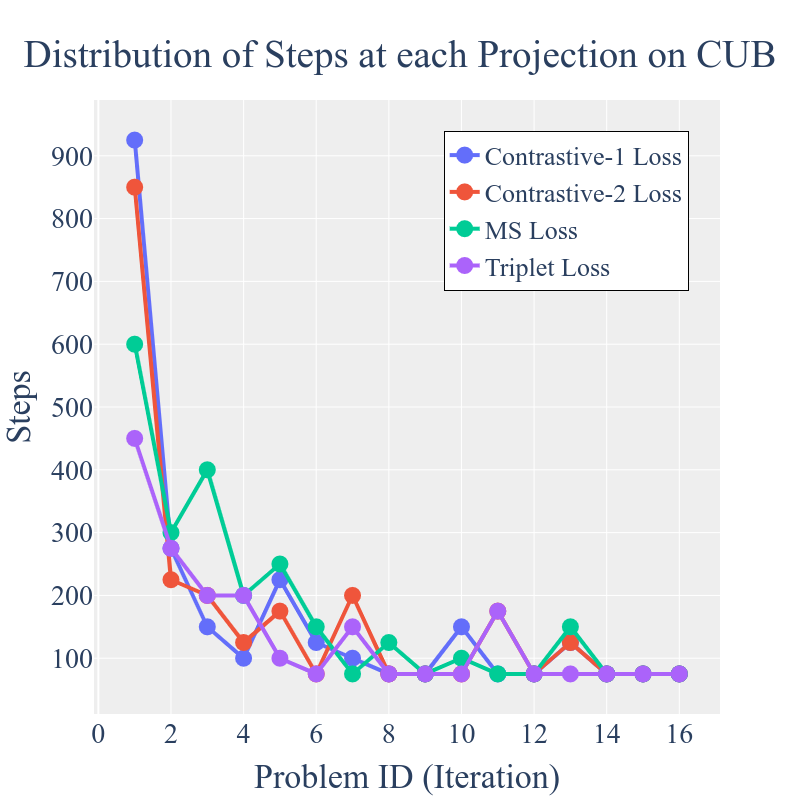}}
  \centerline{\scriptsize{(b)}}

\end{minipage}
\caption{Impact of alternating proxies (a)  and typical distribution of the steps per projection problem (b).}
\label{fig:alternating}
\end{figure}

\textbf{Effect of alternating problems.} We provide results on MNIST in \cref{fig:radius} to show the effect of solving alternating problems instead of single proxy-based DML. We additionally evaluate the baseline losses through solving only a single proxy-DML (\emph{Loss}-Proxy) to show (\cref{fig:alternating}-(a)) that our performance increase is not solely coming from augmentation of proxies in the problem. We clearly observe that alternating proxies helps performance as our formulation suggests. Moreover, we also provide a typical distribution of the steps per proxy-based projection problem in \cref{fig:alternating}-(b) to show that we are not greedy on alternating the proxies just to provide more examples. We do have some relatively small steps, implying the selected proxies are not informative enough to change the geometry of the embedding space.


\begin{figure}[t]
\begin{minipage}{.49\linewidth}

  \centerline{\includegraphics[width=1.0\linewidth,keepaspectratio]{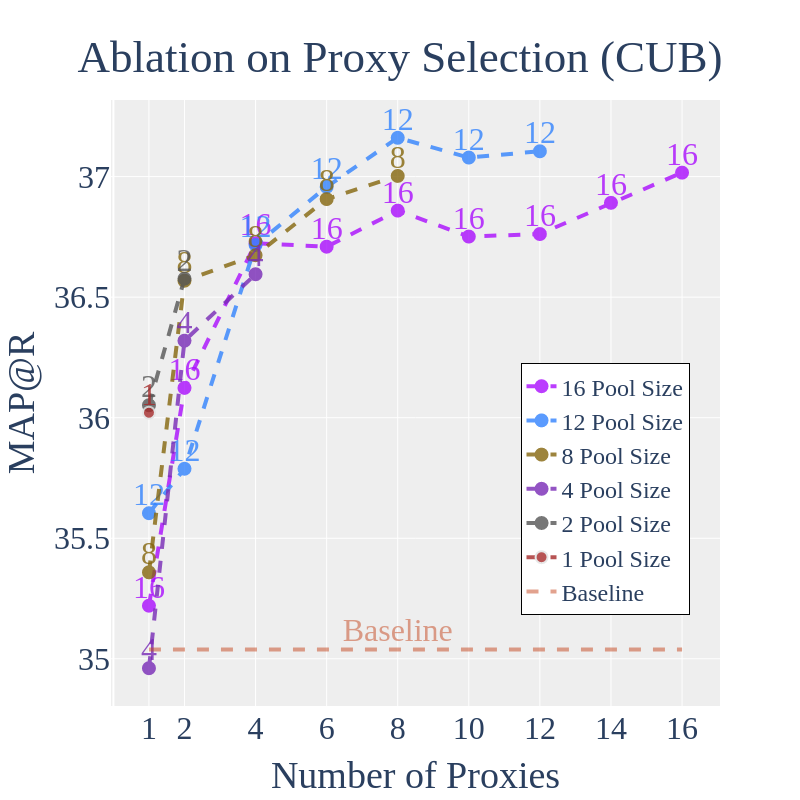}}

\end{minipage}
\begin{minipage}{.49\linewidth}

  \centerline{\includegraphics[width=1.0\linewidth,keepaspectratio]{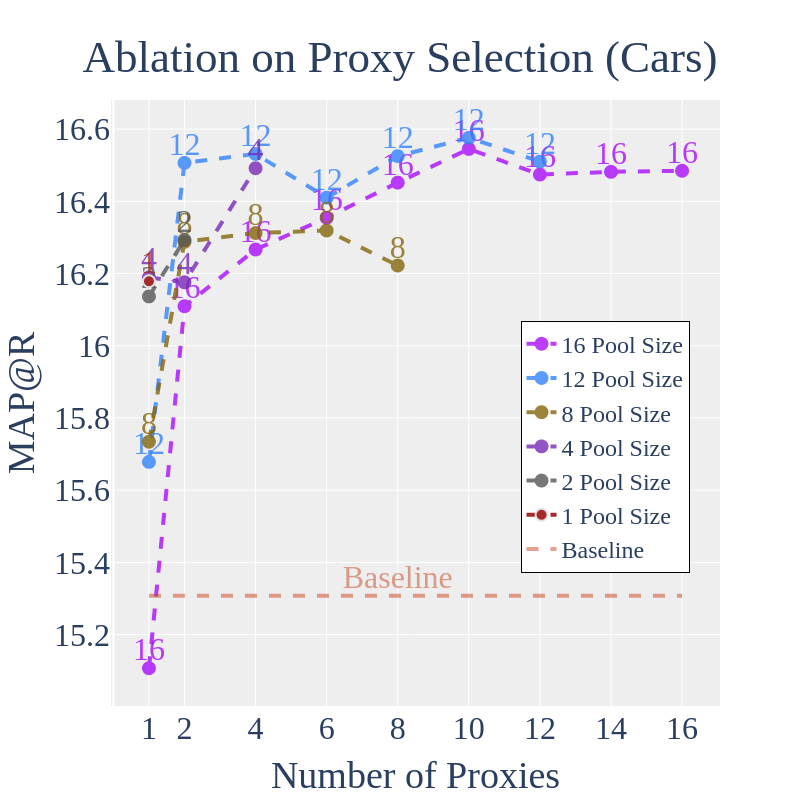}}

\end{minipage}

\caption{Analysis of the relation between the number of proxies and the pool size used for the proxy selection on CUB (left) and Cars (right) dataset with C2-CCP.}
\label{fig:pool_size}
\end{figure}

\textbf{Effect of proxy selection.} We analyze the relation between the number of proxies and the pool size used for the proxy selection with C2-CCP. The related plots are in \cref{fig:pool_size}. We observe that both increasing the number of proxies and the pool size for proxy selection help performance. We interestingly see that for single proxy case, increasing the pool size gives no better results than random selection. Owing to our greedy proxy selection, we do consider the past geometry no earlier than single step. Thus, in the single proxy case, we are prone to oscillate between similar samples for proxy selection. On the other hand, selecting the samples that reduce the covering radius most brings better generalization over random selection. That said, random sampling in proxy selection (\ie, pool size = \#proxy) still works well since random sampling indeed can provide diversity in the samples as well. Such a result supports that the key to our method is alternating the proxies with new samples. As long as we re-initialize the proxies with new samples, we will have some diverse proxies through the iterations. To this end, we use Greedy $K$-Center to pick the samples in a clever way to reduce the covering radius as much as we can (\textit{analogous to mining in batch construction}). 

\section{Conclusion}
Bringing a different perspective to DML formulation, we formulate DML as a chance constrained optimization problem and theoretically show that a contrastive loss based DML objective is a surrogate for the chance constraints. We rigorously convert the initial problem formulation into another form enabling expressing DML as a set intersection problem. The theory suggests a set intersection mechanism yet allows a greedy algorithm via iterative projections. To this end, we also relate the solution of a proxy-based DML approach to one of the supersets to be intersected to obtain the desired solution. As a result, we formally develop a proxy-based method that inherently employs arbitrary number of proxies for better generalization, realizing the mechanism suggested by the theory with a simple, yet effective, algorithm. Supporting our claims, extensive evaluations on 4 DML benchmarks with 4 DML losses showed the effectiveness of our method.



{\small
\bibliographystyle{ieee_fullname}
\bibliography{wacv_2024_paper}
}

\ifx \printOption \printArxiv


\setcounter{section}{0}
\onecolumn
\section*{\Large \centering Supplemental Material for {\large\emph{"Deep Metric Learning with Chance Constraints"}}}
\hypertarget{figmagnified}{\,}

\begin{figure*}[!hb]

\centering
  \centerline{\includegraphics[width=0.8\linewidth,keepaspectratio]{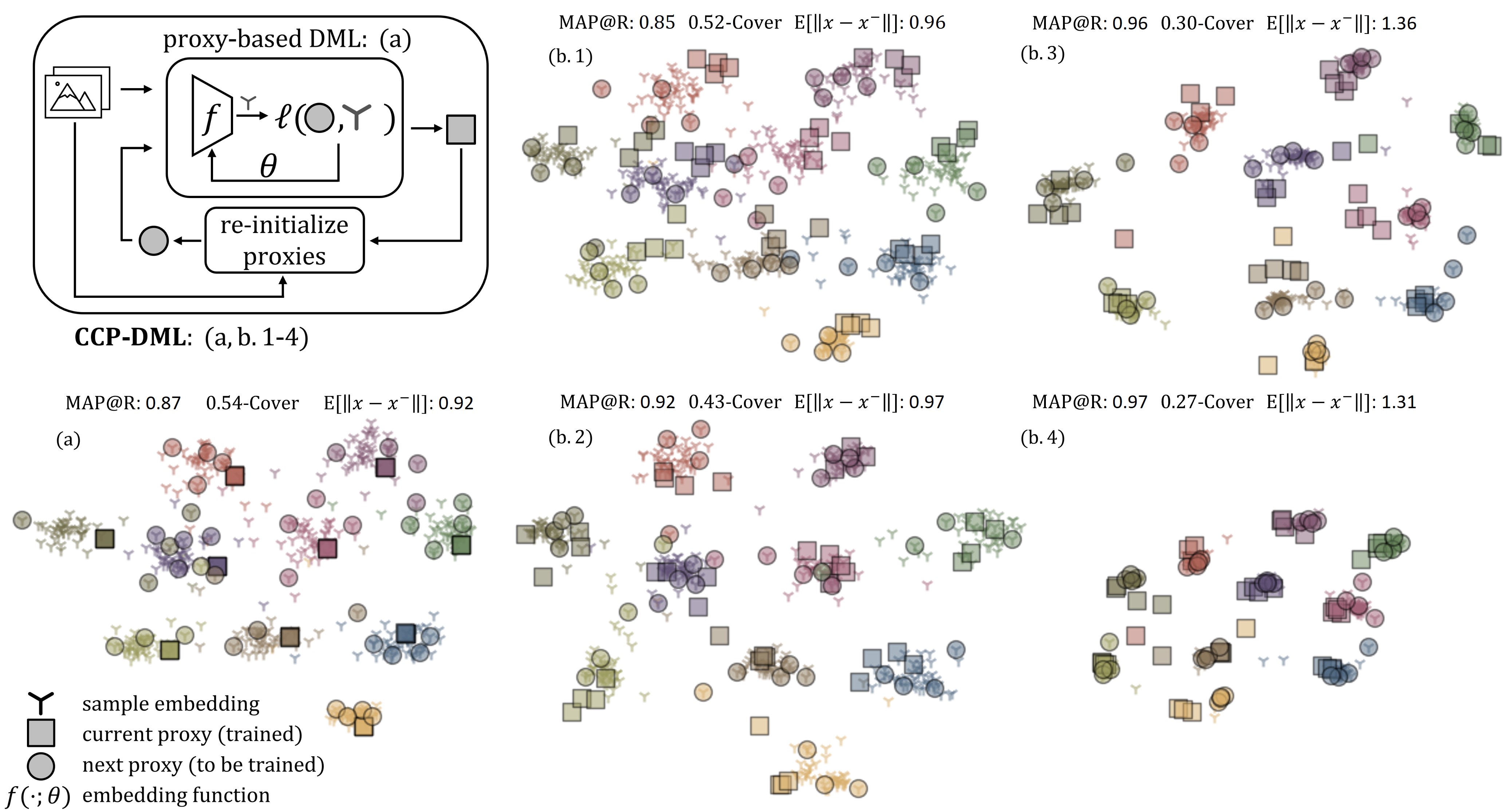}}
  \caption{Illustration of our method (CCP) and the geometry of the embedding space in MNIST dataset: Boxes represent the converged proxies, while circles represent the next proxies resulting from $K$-Center. (a) In proxy-based DML (before our method), proxies coalesce into one. (b) With CCP (through iterations 1-4), diverse proxies are obtained, resulting in a reduced covering radius.}
	\label{fig:radius_bigger}
%
%
%
%
%
%
%

\end{figure*}

\begin{figure*}[!hb]
	\centering
  \centerline{\includegraphics[width=0.68\linewidth,keepaspectratio]{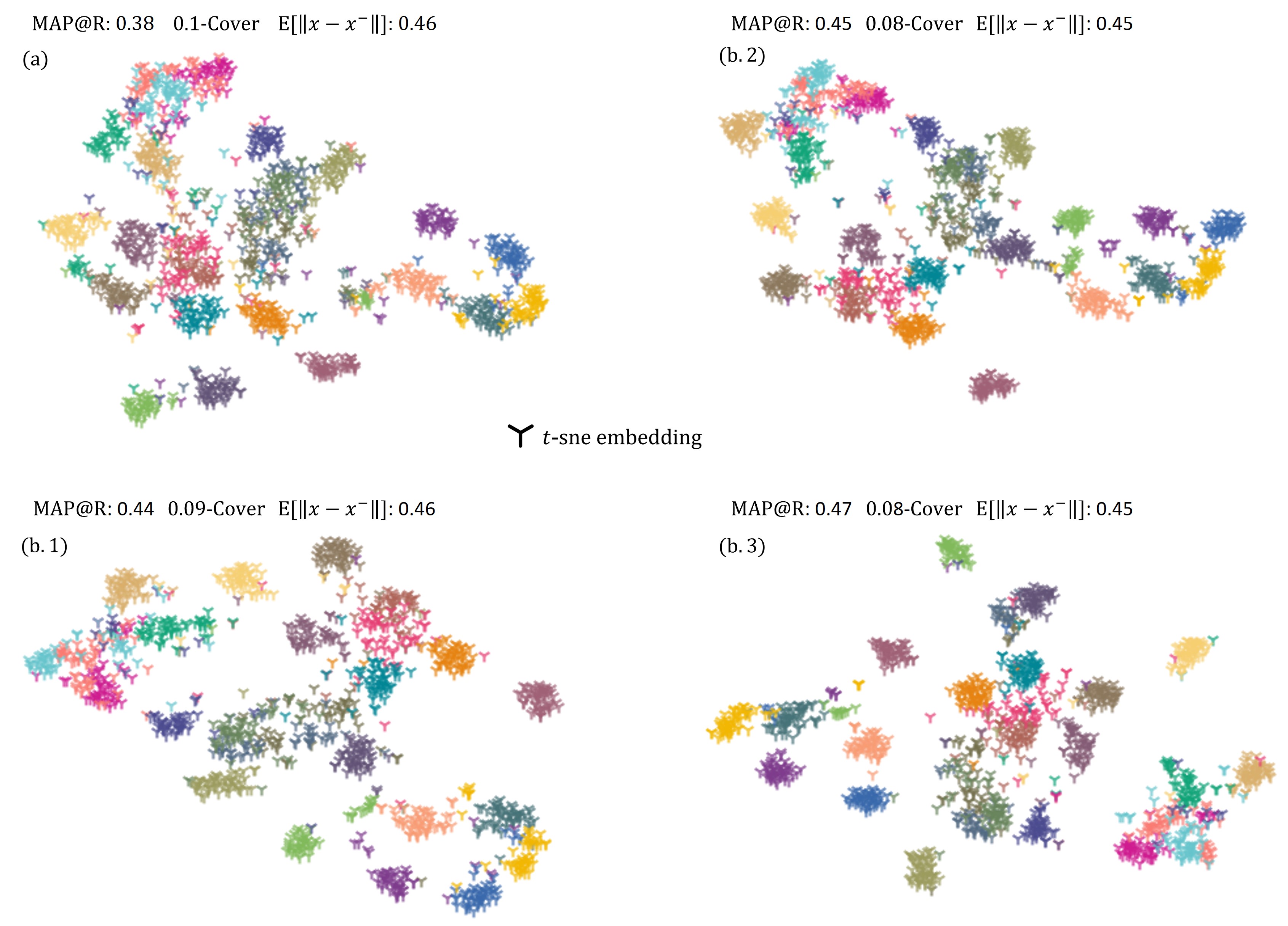}}

\caption{The geometry of the embedding space before, (a), and after, (b), our method (through iterations 1-3), relating how the generalization efforts in training domain transfer to the geometry of test domain on CUB dataset with C2-CCP. We use 2D TSNE embeddings of the validation data in the visualization, in which we report MAP@R, average covering radius and average inter-class pairwise distances.}
\label{fig:radius_test}
\end{figure*}
\clearpage

\begin{table*}[ht]
\centering
\caption{Evaluation on SOP and In-shop for the retrieval task. Red: the overall best. Bold: the loss specific best.}
\label{tab:all_results_bigger_sop_inshop}
\resizebox{\textwidth}{!}{%
\begin{tabular}{@{}ccccccccccccc@{}}
\toprule
                                    & \multicolumn{6}{c}{\textbf{SOP}}                                                                                                                                                                                                                                                        & \multicolumn{6}{c}{\textbf{In-shop}}                                                                                                                                                                                                                                \\ \midrule
                                    & \multicolumn{3}{c|}{\textbf{512D}}                                                                                                         & \multicolumn{3}{c|}{\textbf{128D}}                                                                                                         & \multicolumn{3}{c|}{\textbf{512D}}                                                                                                         & \multicolumn{3}{c}{\textbf{128D}}                                                                                    \\ \cmidrule(l){2-13} 
\textbf{Method}                     & \textbf{P@1}                          & \textbf{P@R}                          & \multicolumn{1}{c|}{\textbf{MAP@R}}                        & \textbf{P@1}                          & \textbf{P@R}                          & \multicolumn{1}{c|}{\textbf{MAP@R}}                        & \textbf{P@1}                          & \textbf{P@R}                          & \multicolumn{1}{c|}{\textbf{MAP@R}}                        & \textbf{P@1}                          & \textbf{P@R}                          & \textbf{MAP@R}                        \\ \midrule
\multicolumn{1}{c|}{C1 \cite{hu2014discriminative}}             & 68.84                                 & 43.28                                 & \multicolumn{1}{c|}{40.25}                                 & 64.96                                 & 39.68                                 & \multicolumn{1}{c|}{36.54}                                 & 80.12                                 & 53.11                                 & \multicolumn{1}{c|}{50.15}                                 & 76.08                                 & 49.61                                 & 46.51                                 \\
\multicolumn{1}{c|}{C1-XBM-L \cite{wang2020cross}}       & 78.68                                 & 54.66                                 & \multicolumn{1}{c|}{51.82}                                 & 75.37                                 & 49.95                        & \multicolumn{1}{c|}{47.39}                                 & 88.39                                 & 61.33                                 & \multicolumn{1}{c|}{58.64}                                 & \textbf{85.75}                        & 58.13                                 & 55.37                                 \\
\multicolumn{1}{c|}{\textbf{C1-CCP-L}}      & {\color[HTML]{FE0000} \textbf{79.53}} & {\color[HTML]{FE0000} \textbf{55.11}} & \multicolumn{1}{c|}{{\color[HTML]{FE0000} \textbf{52.73}}} & {\color[HTML]{FE0000} \textbf{76.24}} & \textbf{50.07}                                 & \multicolumn{1}{c|}{\textbf{48.07}}                        & \textbf{88.52}                        & \textbf{62.54}                        & \multicolumn{1}{c|}{\textbf{59.67}}                        & 85.50                                 & \textbf{58.51}                        & \textbf{55.56}                        \\ \midrule
\multicolumn{1}{c|}{C2 \cite{wu2017sampling}}             & 74.87                                 & 49.88                                 & \multicolumn{1}{c|}{46.94}                                 & 71.15                                 & 45.77                                 & \multicolumn{1}{c|}{42.66}                                 & 86.32                                 & 62.36                                 & \multicolumn{1}{c|}{59.42}                                 & 83.04                                 & 58.27                                 & 55.13                                 \\
\multicolumn{1}{c|}{C2-XBM-L \cite{wang2020cross}}       & 76.66                                 & 51.91                                 & \multicolumn{1}{c|}{49.04}                                 & 73.47                                 & 48.18                                 & \multicolumn{1}{c|}{45.15}                                 & 87.66                                 & 63.50                                 & \multicolumn{1}{c|}{60.64}                                 & 84.58                                 & 59.78                                 & 56.75                                 \\
\multicolumn{1}{c|}{\textbf{C2-CCP-L}}      & \textbf{78.95}                        & \textbf{55.01}                        & \multicolumn{1}{c|}{\textbf{52.19}}                        & \textbf{75.92}                        & {\color[HTML]{FE0000} \textbf{51.14}} & \multicolumn{1}{c|}{{\color[HTML]{FE0000} \textbf{48.18}}} & \textbf{88.52}                        & \textbf{63.94}                        & \multicolumn{1}{c|}{\textbf{61.07}}                        & \textbf{86.11}                        & \textbf{60.16}                        & \textbf{57.24}                        \\ \midrule
\multicolumn{1}{c|}{MS \cite{Wang_2019_CVPR_MS}}             & 72.74                                 & 47.07                                 & \multicolumn{1}{c|}{44.10}                                 & 68.96                                 & 43.25                                 & \multicolumn{1}{c|}{40.18}                                 & 88.37                                 & 63.53                                 & \multicolumn{1}{c|}{60.65}                                 & 85.39                                 & 59.65                                 & 56.61                                 \\
\multicolumn{1}{c|}{\textbf{MS-CCP-L}}      & \textbf{78.96}                        & \textbf{54.71}                        & \multicolumn{1}{c|}{\textbf{51.85}}                        & \textbf{75.80}                        & \textbf{50.48}                        & \multicolumn{1}{c|}{\textbf{47.97}}                        & {\color[HTML]{FE0000} \textbf{90.24}} & \textbf{66.31}                        & \multicolumn{1}{c|}{\textbf{63.59}}                        & {\color[HTML]{FE0000} \textbf{87.10}} & \textbf{61.74}                        & {\color[HTML]{FE0000} \textbf{59.15}} \\ \midrule
\multicolumn{1}{c|}{Triplet \cite{schroff2015facenet}}        & 75.40                                 & 50.13                                 & \multicolumn{1}{c|}{47.03}                                 & 70.41                                 & 44.32                                 & \multicolumn{1}{c|}{41.03}                                 & 86.71                                 & 63.81                                 & \multicolumn{1}{c|}{60.60}                                 & 82.58                                 & 58.74                                 & 55.25                                 \\
\multicolumn{1}{c|}{\textbf{Triplet-CCP-L}} & \textbf{77.09}                        & \textbf{52.42}                        & \multicolumn{1}{c|}{\textbf{49.33}}                        & \textbf{72.21}                        & \textbf{46.38}                        & \multicolumn{1}{c|}{\textbf{43.11}}                        & \textbf{89.44}                        & {\color[HTML]{FE0000} \textbf{67.23}} & \multicolumn{1}{c|}{{\color[HTML]{FE0000} \textbf{64.28}}} & \textbf{86.00}                        & {\color[HTML]{FE0000} \textbf{62.24}} & \textbf{59.04}                        \\ \midrule
\multicolumn{1}{c|}{ProxyAnchor \cite{kim2020proxy}}    & 77.10                                 & 51.95                                 & \multicolumn{1}{c|}{49.01}                                 & 73.86                                 & 47.94                                 & \multicolumn{1}{c|}{44.89}                                 & 88.08                                 & 60.91                                 & \multicolumn{1}{c|}{58.09}                                 & 85.87                                 & 57.80                                 & 54.95                                 \\
\multicolumn{1}{c|}{ProxyNCA++ \cite{teh2020proxynca++}}     & 76.07                                 & 51.17                                 & \multicolumn{1}{c|}{48.20}                                 & 72.89                                 & 47.47                                 & \multicolumn{1}{c|}{44.44}                                 & 87.33                                 & 60.33                                 & \multicolumn{1}{c|}{57.48}                                 & 84.79                                 & 57.33                                 & 54.42                                 \\
\multicolumn{1}{c|}{SoftTriple-S \cite{Qian_2019_ICCV}}   & 78.48                                 & 53.68                                 & \multicolumn{1}{c|}{50.77}                                 & 74.66                                 & 48.79                                 & \multicolumn{1}{c|}{45.75}                                 & 88.37                                 & 62.56                                 & \multicolumn{1}{c|}{59.56}                                 & 85.71                                 & 58.74                                 & 55.68                                \\ 
\multicolumn{1}{c|}{HPL-PA \cite{yang2022hierarchical}}   & 76.97                                 & 51.97                                 & \multicolumn{1}{c|}{49.07}                                 & 73.84                                 & 48.10                                 & \multicolumn{1}{c|}{45.11}                                 & -                                 & -                                 & \multicolumn{1}{c|}{-}                                 & -                                 & -                                 & -                                  \\ \bottomrule
\end{tabular}%
}
\end{table*}
\begin{table*}[!ht]
\centering
\caption{Evaluation on SOP and In-shop for the retrieval task. Red: the overall best. Bold: the loss specific best.}
\label{tab:all_results_bigger_cars_cub}
\resizebox{\textwidth}{!}{%
\begin{tabular}{@{}ccccccccccccc@{}}
\toprule
                                    & \multicolumn{6}{c}{\textbf{CUB}}                                                                                                                                                                                                                                                        & \multicolumn{6}{c}{\textbf{Cars196}}                                                                                                                                                                                                                               \\ \midrule
                                    & \multicolumn{3}{c|}{\textbf{512D}}                                                                                                         & \multicolumn{3}{c|}{\textbf{128D}}                                                                                                         & \multicolumn{3}{c|}{\textbf{512D}}                                                                                                         & \multicolumn{3}{c}{\textbf{128D}}                                                                                    \\ \cmidrule(l){2-13} 
\textbf{Method}                     & \textbf{P@1}                          & \textbf{P@R}                          & \multicolumn{1}{c|}{\textbf{MAP@R}}                        & \textbf{P@1}                          & \textbf{P@R}                          & \multicolumn{1}{c|}{\textbf{MAP@R}}                        & \textbf{P@1}                          & \textbf{P@R}                          & \multicolumn{1}{c|}{\textbf{MAP@R}}                        & \textbf{P@1}                          & \textbf{P@R}                          & \textbf{MAP@R}                        \\ \midrule
\multicolumn{1}{c|}{C1 \cite{hu2014discriminative}}             & 63.67                                 & 33.77                                 & \multicolumn{1}{c|}{23.08}                                 & 56.21                                 & 29.65                                 & \multicolumn{1}{c|}{19.06}                                 & 77.75                                 & 33.69                                 & \multicolumn{1}{c|}{23.50}                                 & 64.17                                 & 26.50                                 & 16.13                                 \\
\multicolumn{1}{c|}{C1-XBM-L \cite{wang2020cross}}       & 65.40                                 & 35.57                                 & \multicolumn{1}{c|}{24.87}                                 & 57.57                                 & 30.42                                 & \multicolumn{1}{c|}{19.84}                                 & 83.68                                 & 37.74                                 & \multicolumn{1}{c|}{27.93}                                 & \textbf{72.13}                        & 28.55                                 & 18.83                                 \\
\multicolumn{1}{c|}{\textbf{C1-CCP-L}}      & \textbf{68.11}                        & \textbf{37.85}                        & \multicolumn{1}{c|}{\textbf{27.11}}                        & \textbf{59.56}                        & \textbf{32.06}                        & \multicolumn{1}{c|}{\textbf{21.27}}                        & \textbf{83.76}                        & \textbf{37.78}                        & \multicolumn{1}{c|}{\textbf{28.32}}                        & 72.05                                 & \textbf{28.74}                        & \textbf{18.96}                        \\ \midrule
\multicolumn{1}{c|}{C2 \cite{wu2017sampling}}             & 67.49                                 & 37.18                                 & \multicolumn{1}{c|}{26.47}                                 & 59.73                                 & 31.86                                 & \multicolumn{1}{c|}{21.01}                                 & 81.04                                 & 34.97                                 & \multicolumn{1}{c|}{24.73}                                 & 69.17                                 & 27.70                                 & 17.22                                 \\
\multicolumn{1}{c|}{C2-XBM-L \cite{wang2020cross}}       & 68.62                                 & 37.53                                 & \multicolumn{1}{c|}{26.83}                                 & 60.18                                 & 32.25                                 & \multicolumn{1}{c|}{21.41}                                 & 82.40                                 & 36.07                                 & \multicolumn{1}{c|}{25.99}                                 & 70.01                                 & 28.49                                 & 18.01                                 \\
\multicolumn{1}{c|}{\textbf{C2-CCP-L}}      & {\color[HTML]{FE0000} \textbf{69.73}} & {\color[HTML]{FE0000} \textbf{38.69}} & \multicolumn{1}{c|}{{\color[HTML]{FE0000} \textbf{28.02}}} & {\color[HTML]{FE0000} \textbf{62.39}} & {\color[HTML]{FE0000} \textbf{33.49}} & \multicolumn{1}{c|}{{\color[HTML]{FE0000} \textbf{22.67}}} & \textbf{82.89}                        & \textbf{36.27}                        & \multicolumn{1}{c|}{\textbf{26.27}}                        & \textbf{72.16}                        & \textbf{28.98}                        & \textbf{18.52}                        \\ \midrule
\multicolumn{1}{c|}{MS \cite{Wang_2019_CVPR_MS}}             & 64.65                                 & 34.84                                 & \multicolumn{1}{c|}{24.15}                                 & 57.24                                 & 30.29                                 & \multicolumn{1}{c|}{19.64}                                 & 80.88                                 & 36.45                                 & \multicolumn{1}{c|}{26.23}                                 & 69.27                                 & 28.93                                 & 18.25                                 \\
\multicolumn{1}{c|}{\textbf{MS-CCP-L}}      & \textbf{68.84}                        & \textbf{38.19}                        & \multicolumn{1}{c|}{\textbf{27.44}}                        & \textbf{61.10}                        & \textbf{33.23}                        & \multicolumn{1}{c|}{\textbf{22.40}}                        & \textbf{86.26} & {\color[HTML]{FE0000} \textbf{38.97}} & \multicolumn{1}{c|}{{\color[HTML]{FE0000} \textbf{29.14}}} & \textbf{74.97}                        & {\color[HTML]{FE0000} \textbf{30.44}} & {\color[HTML]{FE0000} \textbf{19.85}} \\ \midrule
\multicolumn{1}{c|}{Triplet \cite{schroff2015facenet}}        & 64.01                                 & 34.55                                 & \multicolumn{1}{c|}{23.43}                                 & 55.51                                 & 29.38                                 & \multicolumn{1}{c|}{18.51}                                 & 78.44                                 & 33.83                                 & \multicolumn{1}{c|}{23.11}                                 & 64.57                                 & 26.52                                 & 15.68                                 \\
\multicolumn{1}{c|}{\textbf{Triplet-CCP-L}} & \textbf{65.36}                        & \textbf{35.42}                        & \multicolumn{1}{c|}{\textbf{24.53}}                        & \textbf{56.65}                        & \textbf{30.17}                        & \multicolumn{1}{c|}{\textbf{19.31}}                        & \textbf{81.84}                        & \textbf{35.61}                        & \multicolumn{1}{c|}{\textbf{25.21}}                        & \textbf{68.75}                        & \textbf{28.21}                        & \textbf{17.43}                        \\ \midrule
\multicolumn{1}{c|}{ProxyAnchor \cite{kim2020proxy}}    & 68.43                                 & 37.36                                 & \multicolumn{1}{c|}{26.53}                                 & 60.61                                 & 32.36                                 & \multicolumn{1}{c|}{21.48}                                 & 85.29                                 & 37.53                                 & \multicolumn{1}{c|}{27.73}                                 & 75.79 & 29.91                                 & 19.56                                 \\
\multicolumn{1}{c|}{ProxyNCA++ \cite{teh2020proxynca++}}     & 65.48                                 & 35.60                                 & \multicolumn{1}{c|}{24.85}                                 & 58.49                                 & 31.73                                 & \multicolumn{1}{c|}{20.96}                                 & 82.87                                 & 36.56                                 & \multicolumn{1}{c|}{26.34}                                 & 72.45                                 & 29.91                                 & 19.32                                 \\
\multicolumn{1}{c|}{SoftTriple-L \cite{Qian_2019_ICCV}}   & 68.12                                 & 36.98                                 & \multicolumn{1}{c|}{26.02}                                 & 57.94                                 & 30.63                                 & \multicolumn{1}{c|}{19.86}                                 & 84.90                                 & 37.69                                 & \multicolumn{1}{c|}{27.80}                                 & 73.16                                 & 29.60                                 & 19.18                                 \\ 
\multicolumn{1}{c|}{HPL-PA \cite{yang2022hierarchical}}   & 68.25                                 & 37.57                                 & \multicolumn{1}{c|}{26.72}                                 & 61.31                                 & 32.81                                 & \multicolumn{1}{c|}{21.90}                                 & {\color[HTML]{FE0000} \textbf{86.84}}                                 & 38.36                                 & \multicolumn{1}{c|}{28.67}                                 & {\color[HTML]{FE0000} \textbf{76.12}}                                 & 30.13                                 & 19.83                                 \\\bottomrule
\end{tabular}%
}
\end{table*}
\begin{multicols}{2}
\section{Extended Empirical Study for DML}
\subsection{Fair (MLRC) Evaluation on DML Benchmarks}
We follow the procedures proposed in \cite{musgrave2020metric} to provide fair and unbiased evaluation of our method. We provide the full experimental setup details in \cref{sec:setup_supp}. The evaluation results that are summarized in Fig. 3 in the main paper are tabulated in \Cref{tab:all_results_bigger_sop_inshop,tab:all_results_bigger_cars_cub}, which demonstrate the clear superiority of our method, particularly when considering the Mean Average Precision at R (MAP@R) metric.

We should recapitulate that Precision at 1 (P@1), or Recall at 1 (R@1), is a myopic metric for assessing the quality of the embedding space geometry \cite{musgrave2020metric}. Therefore, solely improving P@1 does not necessarily reflect the true order of improvements brought by different methods. As observed in \Cref{tab:all_results_bigger_sop_inshop,tab:all_results_bigger_cars_cub}, methods with similar P@1 (R@1) performances can exhibit more significant differences in MAP@R. Consequently, we firmly believe that comparing MAP@R, rather than P@1 alone, technically provides a more accurate representation of the improvements achieved by our method.


\end{multicols}

\twocolumn

\subsection{Further Ablations}
\label{sec:sup_ablation}

\textbf{Effect of CCP in test domain.} Our \emph{proof of the concept} study in MNIST dataset (\cref{fig:radius_bigger}) empirically shows the implications of our formulation in training domain. It is important to show how such efforts in the training domain are reflected in the test domain since metric learning is expected to be generalized to new classes. We further provide the visualization of the validation data in CUB dataset in \cref{fig:radius_test}. We compute covering radii for 1 to $n$ sample case in k-Center. Namely, we take $k$ samples with minimum cover for $k\in[n]$ where $n$ is the number of samples per class. We then take the average of these radii to compute a representative metric for the covering radius. We observe that solving single proxy-based DML results in relatively poor generalization in the test domain. On the contrary, solving the problem as the set intersection problem with alternating projections improves the embedding geometry (reduced radius without decreased inter-class pairwise distances).

\begin{figure}[!b]
  \centerline{\includegraphics[width=.8\linewidth,keepaspectratio]{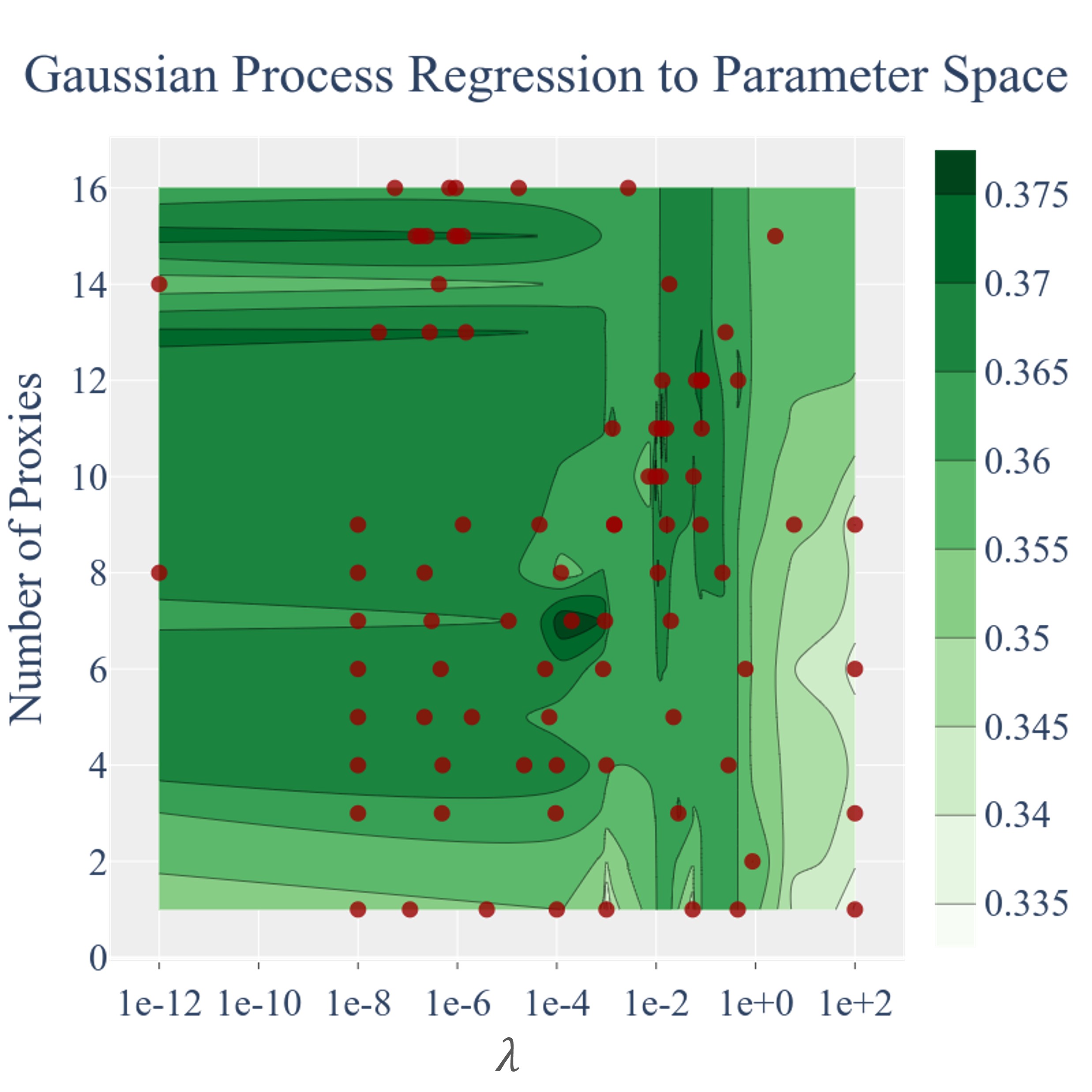}}
\caption{Bayesian search on $\lambda$-$\#proxy$ space}
\label{fig:bayesian}
\end{figure}

\textbf{Hyperparameter search.} Our CCP framework introduces 3 additional hyperparameters to a typical DML problem, which are $\lambda$: regularization weight for the projection objective, \#$proxy$: number of proxies per class, and $b$: pool size for proxy selection in k-Center method. Among those, the selection of \#$proxy$ and $b$ is rather resource dependent and even the setting (\#$proxy=1$, $b=1$) brings performance improvements as we empirically show in Figs. 4 and 5. We expect such a behaviour since CCP mechanism is able to increase the number of proxies inherently. On the other hand, we must analyze the behaviour of CCP with respect to $\lambda$ in order to suggest a proper operation range. To this end, we perform Bayesian search on the $\lambda$-\#$proxy$ space by fixing $b=16$ to see the joint effect of two in CUB with C2-CCP. We provide the results in \cref{fig:bayesian}. We observe that absence of $\lambda$ degrades the performance. Similarly, large values of $\lambda$ causes over-regularization. We obtain interval of $[10^{\shortminus 1}, 10^{\shortminus 5}]$ that works well for $\lambda$. 

For the number of proxies, we observe increasing the proxy per class improves performance. On the other hand, the increase saturates as it can also be observed from Fig. 5. As the result of Bayesian parameter search, we take $\lambda{=}2\cdot10^{\shortminus 4}$ and \#$proxy{=}8$ with pool size $b{=}12$ in our evaluations against other methods for CUB and Cars. For SOP and In-shop, we reduce \#$proxy{=}4$ and $b{=}7$ owing to relatively less number of samples per class in the dataset.

\textbf{Effect of batch size.} Batch size plays important role in DML methods to perform well. Therefore, we analyze the robustness to the batch size especially for the cases where increasing the batch size is prohibitive. We train baseline contrastive loss and CCP contrastive loss for the batch sizes of 16, 32, 64 and 128. The training setup is the same as in the state-of-the-art comparison (\cref{sec:setup_supp}). In each batch we use 4 samples per class. We provide the results in \cref{fig:batch_size}. We observe that baseline contrastive loss has increasing performance as the batch size increases whereas our method's performance with small batch size is on par with the large batch size. Thus, CPP has reduced batch size complexity.

\begin{figure}[!h]
\centering
\begin{minipage}{.49\linewidth}

  \centerline{\includegraphics[width=1.0\linewidth,keepaspectratio]{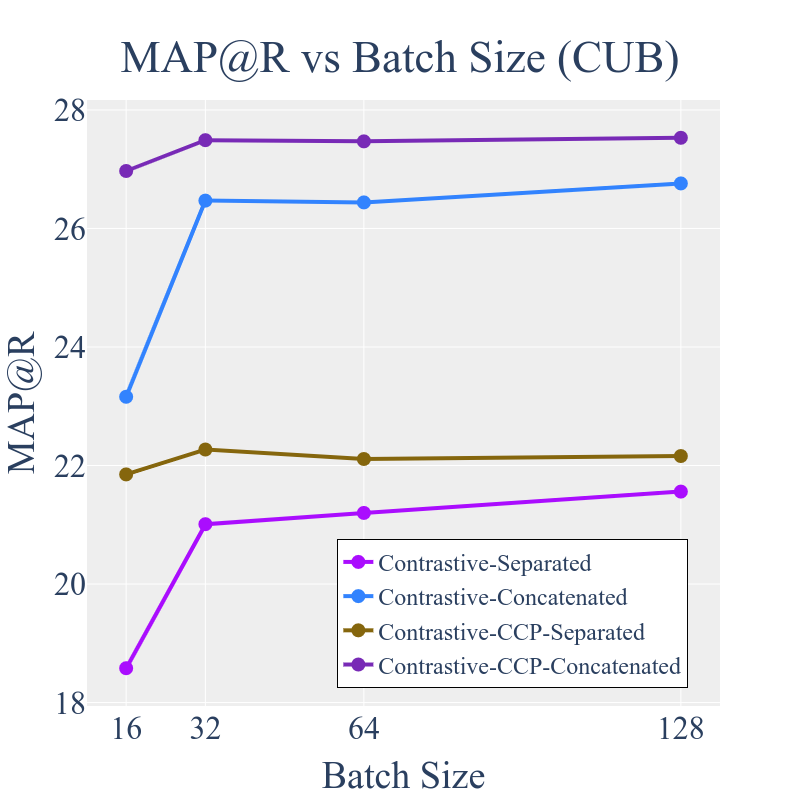}}

\end{minipage}
\begin{minipage}{.49\linewidth}

  \centerline{\includegraphics[width=1.0\linewidth,keepaspectratio]{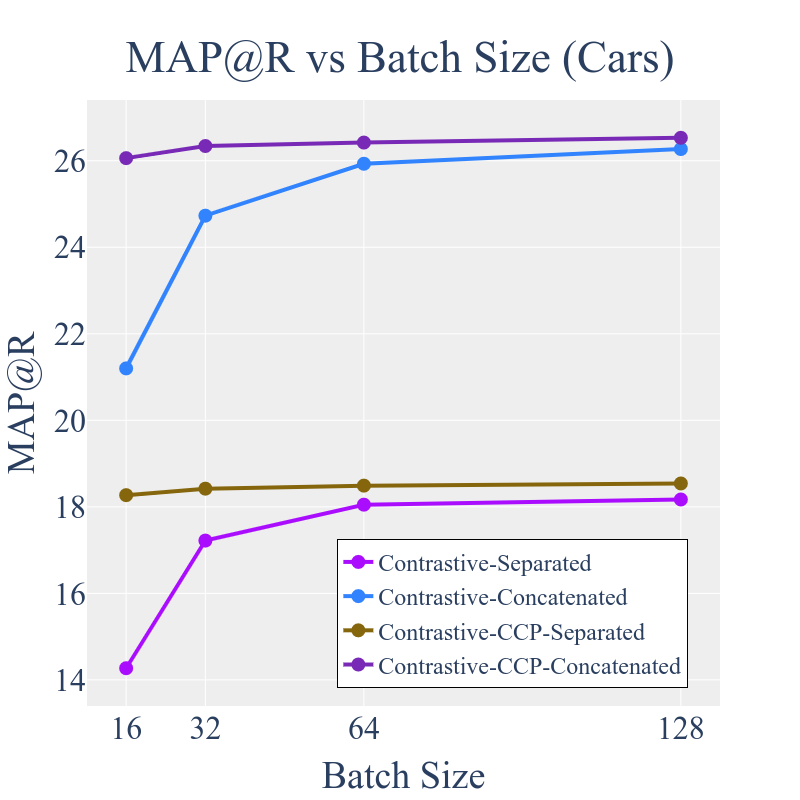}}

\end{minipage}

\caption{Analysis of batch size dependence of the performance on CUB (left) and Cars (right) dataset with C2-CCP.}
\label{fig:batch_size}

\end{figure}

\textbf{Computational analysis.} Our method outlined in Algorithm 1 puts marginal computation and memory overhead on top of the baseline approaches. 

For the computation, we have proxy initialization and weight update steps at the beginning of the each problem instance. In overall, in our system with RTX 2080 Ti GPU and i7 CPU, that additional computation adds on the average ~5-10 ms per step (batch update). In particular, for batch size of 32, we typically have rate of 105 ms/batch with Contrastive-CCP whereas vanilla has 97 ms/batch rate. In In-shop and SOP dataset, we have the same rates however for CCP, we have 200 to 400 ms computation overhead due to sampling for proxy initialization. We do not have such overhead in Cars and CUB owing to the much less number of classes. With that being said, we have such 400 ms overhead in In-shop and SOP only at the beginning of new problem instance, which has no significant
%
\begin{wraptable}[14]{r}[-5pt]{0.5\linewidth}
\vspace{-0.75\intextsep}
\centering
\caption{Total steps of training in SOP and In-shop}
\label{tab:steps2convergence}
\resizebox{\linewidth}{!}{%
\begin{tabular}{@{}ccc@{}}
\toprule
\textbf{Method} & \textbf{SOP} & \textbf{In-shop} \\ \midrule
C2              & 62K          & 114K             \\
C2-XBM          & 81K          & 93K              \\
C2-CCP          & 69K          & 127K             \\
MS              & 67K          & 98K              \\
MS-CCP          & 91K          & 131K             \\
Triplet         & 93K          & 73K              \\
Triplet-CCP     & 124K         & 115K             \\
ProxyAnchor     & 54K          & 87K              \\
ProxyNCA++      & 88K          & 103K             \\
SoftTriple      & 48K          & 82K              \\ \bottomrule
\end{tabular}%
}
\end{wraptable}
\!\!effect in long run thanks to rather infrequent happening of proxy re-initialization. Due to alternating problems, our method takes more steps to converge than their baseline counterparts. We provide the optimization steps per proxy-based problem instance for several losses in Fig. 4 from which relative convergence can be compared owing to each problem instance being a proxy-based method. Nevertheless, we also provide \cref{tab:steps2convergence} to compare the convergence of the methods for SOP and In-shop datasets. The reported numbers are the rounded averages of the 4 models. We observe  10\%-35\% increase in the optimization steps for the pairwise losses.

For the memory, we store the weighs of the previously converged model in the memory as well as the variables for proxies. For the model, approximately 40-45 mb additional GPU memory is used and for the proxies 16.6 mb and 5.9 mb memory is used in SOP and In-shop dataset (75 kb in CUB and Cars). 

In summary, CCP brings $\approx 8\%$ increase in back-propagation computation time and only $\approx 60$ MB increase in memory for the largest model. CCP takes $10\%\shortminus35\%$ more steps to converge than their baseline counterparts due to alternating problems. On the other hand, our method with small batch size performs on par with the large batch size thanks to alternating proxies (\cref{fig:batch_size}). To this manner, marginal increase in computation is seemingly a fair trade-off in improving the performance along with robustness to batch size. 

\section{Empirical Study Details}
\label{sec:sup_emp_details}

In the following sections, we outline the complete details of our experimental setup, enabling easy reproducibility.

We use our own framework implemented in Tensorflow \cite{abadi2016tensorflow} library in the experiments.

\textbf{Fairness in evaluation.} Independent works \cite{roth2020revisiting, musgrave2020metric, fehervari2019unbiased} reveal that conventional training and evaluation procedures in DML may fail to properly assess the true order of performance that the methods bring. The consensus for unbiased comparability is evaluation of the methods with their best version under the same experimental settings unless the compared methods demand any particular architecture or experimental setup. Our empirical study is completely aligned with the literature’s claims for unbiased evaluation.

\textbf{Reproducibility.} We provide full detail of our experimental setup and recapitulate the implementation details for the sake of complete transparency and reproducibility. Code is available at: \href{\codeurl}{CCP-DML Framework}

\subsection{Experimental Setup}
\label{sec:setup_supp}

\textbf{Datasets.} We perform our experiments on 4 widely-used benchmark datasets: Stanford Online Products (SOP) \cite{oh2016deep}, In-shop \cite{liu2016deepfashion}, Cars196 \cite{krause2014submodular} and, CUB-200-2011 (CUB) \cite{wah2011caltech}.

\textbf{SOP} \cite{oh2016deep} has 22,634 classes with 120,053 product images. The first 11,318 classes (59,551 images) are split for training and the other 11,316 (60,502 images) classes are used for testing.

\textbf{In-shop} has 7,986 classes with 72,712 images. We use 3,997 classes with 25,882 images as the training set. For the evaluation, we use 14,218 images of 3,985 classes as the query and 12,612 images of 3,985 classes as the gallery set.

\textbf{Cars196} contains 196 classes with 16,185 images. The first 98 classes (8,054 images) are used for training and remaining 98 classes (8,131 images) are reserved for testing.

\textbf{CUB-200-2011} dataset consists of 200 classes with 11,788 images. The first 100 classes (5,864 images) are split for training, the rest (5,924 images) is used for testing.

\textbf{Training Splits.} \hypertarget{code}{We} split datasets into disjoint training, validation and test sets according to \cite{musgrave2020metric}. In particular, we partition $\nicefrac{50\%}{50\%}$ for training and test, and further split training data to 4 partitions where 4 models are to be trained by exploiting $\nicefrac{1}{4}$ as validation while training on $\nicefrac{3}{4}$. For the ablation studies, we split training set into 3 splits instead of 1 and train a single model on the $\nicefrac{2}{3}$ of the set while using $\nicefrac{1}{3}$ for the validation.

\textbf{Data augmentation} follows \cite{musgrave2020metric}. During training, we  resize each image so that its shorter side has length 256, then make a random crop between 40 and 256, and aspect ratio between $\nicefrac{3}{4}$ and $\nicefrac{4}{3}$. We resize the resultant image to $227\sxtimes 227$ and apply random horizontal flip with $50\%$ probability. During evaluation, images are resized to 256 and then center cropped to $227\sxtimes 227$.

\phantomsection
\label{sec:metrics}
\textbf{Evaluation metrics.} \hypertarget{metric}{We} consider precision at 1 (P@1), precision (P@R) and mean average precision (MAP@R) at R where R is defined for each query\footnote{A query is an image for which similar images are to be retrieved, and the references are the images in the database.} and is the total number of true references as the query. Among those, MAP@R performance metric is shown to better reflect the geometry of the embedding space and to be less noisy as the evaluation metric \cite{musgrave2020metric}. Thus, we use MAP@R to monitor training.

\textit{P@1:} Find the nearest reference to the query. The score for that query is 1 if the reference is of the same class, 0 otherwise. Average over all queries gives P@1 metric.

\textit{P@R:} For a query $i$, find $R_i$ nearest references to the query and let $r_i$ be the number of true references in those $R_i$-neighbourhood. The score for that query is \mbox{$\text{P@R}_i=\nicefrac{r_i}{R_i}$}. Average over all queries gives P@R metric, \ie, \mbox{$\text{P@R} =\tfrac{1}{n}\!\sum\limits_{i\in[n]}\!\text{P@R}_i$}, where $n$ is the number of queries.

\textit{MAP@R:} We define \mbox{$\text{MAP@R}_i \coloneqq \tfrac{1}{R_i}\sum\limits_{j\in[R_i]} P(j)$} for a query $i$, where $P(j)=\text{P@j}$ if $j^\text{th}$ retrieval is correct or 0 otherwise. Average over all queries gives MAP@R metric, \ie, $\text{MAP@R} = \tfrac{1}{n}\sum\limits_{i\in[n]}\text{MAP@R}_i$, where $n$ is the number of queries. 

\textbf{Training procedure.} For the optimization procedure, we use Adam \cite{kingma2014adam} optimizer for mini-batch gradient descent with a mini-batch size of 32 (4 samples per class), $10^{\shortminus 5}$ learning rate, $10^{\shortminus 4}$ weight decay, default moment parameters, $\beta_1{=}.9$ and $\beta_2{=}.99$. We evaluate validation MAP@R for every 25 steps of training in CUB and Cars196, for 250 steps in SOP and In-shop. We stop training if no improvement is observed for 60 steps and recover the parameters with the best validation performance. Following \cite{musgrave2020metric}, we train 4 models for each $\nicefrac{3}{4}$ partition of the train set. For the ablation studies, we train a single model on the $\nicefrac{2}{3}$ partition.

\textbf{Embedding vectors.} Embedding dimension is fixed to 128. During training and evaluation, the embedding vectors are L2 normalized using the transformation proposed in Section 4.4. We follow the evaluation method proposed in \cite{musgrave2020metric} and produce two results: $i)$ Average performance (128 dimensional) of 4-fold models and $ii)$ Ensemble performance (concatenated 512 dimensional) of 4-fold models where the embedding vector is obtained by concatenated 128D vectors of the individual models.

\textbf{Losses with CCP.} We evaluate our method with \textit{C1-CCP}: Contrastive loss \cite{hadsell2006dimensionality}, \textit{C2-CCP}: Contrastive loss with positive margin \cite{wu2017sampling}, \textit{MS-CCP}: Multi-similarity (MS) loss \cite{Wang_2019_CVPR_MS}, \textit{Triplet-CCP}: Triplet loss \cite{schroff2015facenet}. We should note that ProxyAnchor \cite{kim2020proxy} is indeed proxy-based MS loss except for missing a margin term. Similarly, ProxyNCA \cite{teh2020proxynca++} is $\log\Sigma\exp$-approximation of proxy-based Triplet with hard negative mining and for single proxy case SoftTriple \cite{Qian_2019_ICCV} is equivalent to ProxyNCA. Therefore, our experiments cover wide range of the DML losses.

\textbf{Compared methods.} We compare our method against proxy-based SoftTriple \cite{Qian_2019_ICCV}, ProxyAnchor \cite{kim2020proxy} and ProxyNCA++ \cite{teh2020proxynca++} methods as well as XBM \cite{wang2020cross}.

\textbf{Fairness.} We note that like the compared methods (\ie, loss functions, proxy-based methods), our method’s improvement claims do not demand any particular architecture or experimental setup. Therefore, to evaluate the improvements purely coming from the proposed ideas, we implemented the best version of the compared methods in our framework and evaluate on the same architecture and experimental settings. In this manner, we stick to BN-Inception with global average pooling architecture to directly compare our method with the benchmarked losses in \cite{musgrave2020metric}. To eliminate any framework related performance differences, we re-implemented the methods within our framework and produce the consistent results with \cite{musgrave2020metric}. Our experimental setting is fair and unbiased since:
\\
$\mathbf{i)}$ The compared methods are either invented loss functions or proxy-based approaches, which do not demand a particular setting to show the effectiveness of the proposed ideas.
\\
$\mathbf{ii)}$ We use the same experimental setting for each method (\eg, image size, architecture, embedding size, batch size, data augmentation).
\\
$\mathbf{iii)}$ We implement and re-evaluate all the compared methods on our framework (\ie, train and evaluate).
\\
$\mathbf{iv)}$ We reproduce consistent results reported in \cite{musgrave2020metric} to eliminate any framework related performance bias.
\\
$\mathbf{v)}$ We use the same train and test split as the conventional methods, but we do not exploit test data during training. We use $\nicefrac{1}{4}$ split of train data for the validation set.

\textbf{Hyperparameters.} For the hyperparameter selection, we exploit the recent work \cite{musgrave2020metric} that has performed parameter search via Bayesian optimization on variety of losses. We further experiment the suggested parameters from the original papers and official implementations. We pick the best performing parameters. We perform no further parameter tuning for the loss parameters when applied to our method to purely examine the effectiveness of our method. 

\textit{C1}: We adopted XBM's official implementation for fair comparison. We use 0.5 margin for all datasets.

\textit{C2}: C2 has two parameters, $(m^{+}, m^{-})$: positive margin, $m^{+}$, and negative margin. We set $(m^{+}, m^{-})$ to $(0, 0.3841)$, $(0.2652, 0.5409)$, $(0.2858, 0.5130)$, $(0.2858, 0.5130)$ for CUB, Cars196, In-shop and SOP, respectively.

\textit{Triplet}: We set its margin to 0.0961, 0.1190, 0.0451, 0.0451 for CUB, Cars196, In-shop and SOP, respectively.

\textit{MS}: We set its parameters $(\alpha, \beta, \lambda)$  to $(2, 40, 0.5)$, $(14.35, 75.83, 0.66)$, $(8.49, 57.38, 0.41)$, $(2, 40, 0.5)$ for CUB, Cars196, In-shop and SOP, respectively.

\textit{ProxyAnchor}: We set its two paremeters $(\delta, \alpha)$ to $(0.1, 32)$ for all datasets. We use 1 sample per class in batch setting (\ie, 32 classes with 1 samples per batch), we perform 1 epoch warm-up training of the embedding layer, and we apply learning rate multiplier of 100 for the proxies.

\textit{ProxyNCA++}: We set its temperature parameter to 0.1 for all datasets. We use 1 sample per class in batch setting (\ie, 32 classes per batch), we perform 1 epoch warm-up training of the embedding layer, and we apply learning rate multiplier of 100 for the proxies during training.

\textit{SoftTriple}: SoftTriple has 4 parameters  $\lambda$, $\gamma$, $\tau$, and $\delta$. We set $(\lambda, \gamma, \tau, \delta)$ to $($20, 0.1, 0.2, 0.01$)$, $($17.69, 19.18, 0.0669, 0.3588$)$, $(20, 0.1, 0.2, 0.01)$, $(100, 47.9, 0.2, 0.3145)$ for CUB, Cars196, In-shop and SOP, respectively. We use 1 sample per class in batch setting (\ie, 32 classes with 1 samples per batch), we perform 1 epoch warm-up training of the embedding layer, and we apply learning rate multiplier of 100 for the proxies during training.

\textit{XBM}: We evaluate XBM with C1 and C2 since in the original paper, contrastive loss is reported to be the best performing baseline with XBM. We set the memory size of XBM to the total number of proxies (\ie, $proxy\_per\_class\times\#classes$) to compare the methodology by disentangling the effect of proxy number. With that being said, we also evaluate XBM with the memory sizes suggested in the original paper. In this manner we use two memory sizes for XBM for each dataset: $(S, L)$ where $S$ and $L$ denote the number of batches in the memory. For CUB and Cars196, CCP uses 1(8) proxies per class for $S(L)$ . Thus, we set $(S, L)$ to $(3, 25)$ for CUB and Cars196. For In-shop and SOP, CCP uses 1(4) proxies per class for $S(L)$. Thus, we set $(S, L)$ to $(100, 400), (400, 1400)$ for In-shop and SOP, respectively. We perform 1K steps of training with the baseline loss prior to integrate XBM loss in order to ensure \textit{slow drift} assumption.

\textit{CCP}: For the hyperparameters of our method, we use 8 proxies per class and $\lambda{=}2\,10^{\shortminus 4}$ for CUB and Cars datasets, as the result of the parameter search; and use pool size, $b{=}12$, for greedy k-Center method. We select pool size based on our empirical studies on the effect pool size and number of proxies. Due to computation limitations, we use 4 proxy per class, $\lambda{=}2\,10^{\shortminus 4}$ and $b{=7}$ for SOP and In-shop dataset. We perform no warm-up or do not use learning rate multiplier for the proxies.

\newpage

\appendix
\onecolumn
\appendix
\section{Appendix}
\label{sec:all_proofs}
\subsection{Proof for Lemma 4.1}

\begin{customlemma}{4.1}\label{lemma41}
Generalized contrastive loss defined as $\ell(z_i,z_j;\theta)\coloneqq(y_{ij}(\Vert x_i \shortminus x_j\Vert_{f_\theta} - \beta) + \alpha)_+$ is $\sqrt{2}\omega^L$-Lipschitz in $x_i$ and $x_j$ for all $y_i,y_j,\theta$ for the embedding function $f(\cdot;\theta)$ being $L$-layer CNN (with ReLU, max-pool, average-pool) with a fully connected layer at the end, where $\omega$ is the maximum sum of the input weights per neuron.
\end{customlemma}

\begin{proof}
We first show that $f(x;\theta)$ is Lipschitz continuous.

We consider $x{\in}\Real^d$ as an input to a layer and $\hat{x}{\in}\Real^{d^\prime}$ as the corresponding output. We express $i^{th}$ component of $\hat{x}$ as $ \hat{x}_i = \textstyle\sum_j w_{i,j}x_{s_i(j)}$ where $s_i =\lbrace s_i(j)\in [d] \rbrace$ is the set of components contributing to $\hat{x}_i$ and $w_{i,j}{\in}\theta$ is the layer weights. For instance, for a fully connected layer $s_i(j)=j$; for a $3\sxtimes 3$ convolutional layer, $s_i$ corresponds to $3\sxtimes 3$ window of depth $\#channels$ centered at $i$. We now consider two inputs $x,x^\prime$ and their outputs $\hat{x},\hat{x}^\prime$. We write:
\begin{equation}
\nonumber
\begin{split}
\dfrac{\Vert \hat{x} - \hat{x}^\prime\Vert_2^2}{\Vert x - x^\prime\Vert_2^2} &= 
\dfrac{\textstyle\sum_{i\in[d^\prime]}\vert\hat{x}_i - \hat{x}^\prime_i\vert^2}{\Vert x - x^\prime\Vert_2^2} =
 \dfrac{\textstyle\sum_{i\in[d^\prime]}\vert \textstyle\sum_j w_{i,j}x_{s_i(j)} - \textstyle\sum_j w_{i,j}x^\prime_{s_i(j)} \vert^2}{\Vert x - x^\prime\Vert_2^2} \\
&\leqslant \dfrac{\textstyle\sum_{i\in[d^\prime]} \textstyle\sum_j \vert w_{i,j}\vert^2\vert x_{s_i(j)} - x^\prime_{s_i(j)} \vert^2}{\Vert x - x^\prime\Vert_2^2}
\end{split}
\end{equation}
Rearranging terms, we express:
$$\textstyle\sum_{i\in[d^\prime]} \textstyle\sum_j \vert w_{i,j}\vert^2\vert x_{s_i(j)} - x^\prime_{s_i(j)} \vert^2 = 
\textstyle\sum\limits_{k\in[d]} \textstyle\sum\limits_{i,j:s_i(j)=k} \!\!\!\!\!\!\! \vert w_{i,j}\vert^2\vert x_k - x^\prime_k \vert^2$$
If $\sum\limits_{i,j:s_i(j)=k} \!\!\!\!\!\!\!\!\! \vert  w_{i,j}\vert\leqslant \omega$ for all $k$ and for all layers, \ie, the absolute sum of the input weights per neuron is bounded by $\omega$, we can write $\textstyle\sum\limits_{k\in[d]} \textstyle\sum\limits_{i,j:s_i(j)=k} \!\!\!\!\!\!\! \vert w_{i,j}\vert^2\vert x_k - x^\prime_k \vert^2 \leqslant \omega^2\!\!\textstyle\sum\limits_{k\in[d]}\!\!\vert x_k - x^\prime_k \vert^2\leqslant \omega^2\Vert x-x^\prime \Vert_2^2 $, hence,
$$ \dfrac{\Vert \hat{x} - \hat{x}^\prime\Vert_2}{\Vert x - x^\prime\Vert_2} \leqslant \omega.$$
For max-pooling and average-pooling layers, the inequality holds with $\omega=1$; since, we can express max-pooling as a convolution where only one weight is 1 and the rest is 0; and similarly, we can express average-pooling as a convolution where the weights sum up to 1.

For ReLU activation, we consider the fact that
$\vert \max\lbrace 0, u\rbrace - \max\lbrace 0, v\rbrace\vert \leqslant \vert u - v\vert$ to write:
$$ \dfrac{\Vert ReLU(x) - ReLU(x^\prime) \Vert_2}{\Vert x - x^\prime \Vert_2} \leqslant 1. $$

Therefore, $L$-layer CNN $f(x;\theta)$ is $\omega^L$-Lipschitz.

We now consider $\ell(z_i,z_j;\theta) = \max\lbrace 0, y_{ij}(\Vert f(x_i;\theta)-f(x_j;\theta)\Vert_2 - \beta) + \alpha\rbrace$ as $g(h(f(x_i;\theta), f(x_j;\theta)))$ where $g(h)=\max\lbrace 0,y_{ij}(h-\beta)+\alpha\rbrace$ is 1-Lipschitz, and $h(f, f^\prime) = \Vert f - f^\prime \Vert_2$ is $\sqrt{2}$-Lipschitz and 1-Lipschitz in $f$ for fixed $f^\prime$. Thus, for $y_i,y_j,\theta$ fixed, $\ell(z_i,z_j;\theta)\coloneqq(y_{ij}(\Vert x_i \shortminus x_j\Vert_{f_\theta} - \beta) + \alpha)_+$ is $\omega^L$-Lipschitz in $x_i$ and in $x_j$; and $\sqrt{2}\omega^L$-Lipschitz in both, for all $y_i,y_j,\theta$.
\end{proof}

Note that it is easy to show that the normalization proposed in Section 4.4:
\begin{equation}
\nonumber
\hat{v} =
\begin{cases}
v & \text{for } \Vert v \Vert_2 \leqslant 1 \\ 
\nicefrac{v}{\Vert v \Vert_2}  & \text{for } \Vert v \Vert_2 \geqslant 1 
\end{cases}
\end{equation}
is 2-Lipschitz. Therefore, our loss is still Lipschitz continuous with normalized embeddings in our framework.

\subsection{Proof for Proposition 4.1}

\begin{customprop}{4.1}\label{prop41}
Given $\mathcal{S}{=}\lbrace z_i\rbrace_{i\in[m]} {\overset{i.i.d.}{\sim}} p_\mathcal{Z}$ such that $\forall k{\in}\mathcal{Y}$ $\lbrace x_i{\mid} y_i{=}k\rbrace$ is $\delta_\mathcal{S}$-cover\footnote{$\mathcal{S}\subset \mathcal{S}^\prime$ is $\delta_\mathcal{S}$-cover of $\mathcal{S}^\prime$ if $\forall z^\prime \in \mathcal{S}^\prime$, $\exists z\in \mathcal{S}$ such that $\Vert z-z^\prime\Vert_2 \leqslant \delta_\mathcal{S}$.} of $\mathcal{X}$, $\ell(z_i,z_j;\theta)$ is $\zeta$-Lipschitz in $x_i, x_j$ for all $y_i$, $y_j$ and $\theta$, and bounded by $L$; then with probability at least $1-\gamma$,
$$\Big\vert\E{\ell(z_i,z_j;\theta)}{z_i,z_j} - \tfrac{1}{m}\!\textstyle\sum\limits_{i\in[m]}\E{\ell(z_i,z_j;\theta)}{z_j}\Big\vert\leqslant \mathcal{O}(\zeta\,\delta_\mathcal{S}) + \mathcal{O}(L\,\sqrt{\nicefrac{\log\tfrac{1}{\gamma}}{m}}).$$
\end{customprop}

\begin{proof}
We start with defining $\hat{\mathcal{L}}(z;\theta)\coloneqq \E{\ell(z, z^\prime;\theta)}{z^\prime\sim p_\mathcal{Z}}$. Note that
\begin{equation}
\begin{split}
\nonumber
\Vert\hat{\mathcal{L}}(z_1;\theta)-\hat{\mathcal{L}}(z_2;\theta)\Vert_2 &=
\vert\E{\ell(z_1, z^\prime;\theta)}{z^\prime\sim p_\mathcal{Z}}-\E{\ell(z_2, z^\prime;\theta)}{z^\prime\sim p_\mathcal{Z}}\vert\\
&\leqslant
\E{\vert\ell(z_1, z^\prime;\theta)-\ell(z_2, z^\prime;\theta)\vert}{z^\prime\sim p_\mathcal{Z}}.
\end{split}
\end{equation}
Therefore, $\ell(z,z^\prime;\theta)$ being $\zeta$-Lipschitz in $x$ for fixed $x^\prime$, $y, y^\prime$ and $\theta$, and bounded by $L$ implies $\hat{\mathcal{L}}(z;\theta)$ is also $\zeta$-Lipschitz in $x$ for all $y$, $\theta$ and bounded by $L$. Hence, we have
$$ \vert \hat{\mathcal{L}}(z_i;\theta) - \hat{\mathcal{L}}(z;\theta) \vert\leqslant \zeta\,\delta_{\mathcal{S}} \quad \forall z_i,z: z_i\in \mathcal{S}, z\in\mathcal{Z}, \Vert z_i - z\Vert_2\leqslant \delta_{\mathcal{S}} $$
From Theorem 14 of \cite{xu2012robustness}, we can partition $\mathcal{Z}$ into $K=\min_t\lbrace \vert t \vert: t \text{ is } \tfrac{\delta_{\mathcal{S}}}{2}\text{-cover of } \mathcal{Z} \rbrace$ disjoint sets, denoted as $\lbrace \mathcal{R}_i\rbrace_{i\in[K]}$, such that $\forall i: z_i\in \delta_{\mathcal{S}}$; both $z_i,z$ being $\in \mathcal{R}_i$ implies $\vert \hat{\mathcal{L}}(z_i;\theta) - \hat{\mathcal{L}}(z;\theta) \vert\leqslant \zeta\,\delta_\mathcal{S}$. Hence, from Theorem 3 of \cite{xu2012robustness}, with probability at least $1-\gamma$, we have:
\begin{equation}
\begin{split}
\nonumber
\Big\vert \E{\ell(z,z^\prime;\theta)}{z,z^\prime \sim p_\mathcal{Z}} - \tfrac{1}{m} \textstyle\sum\limits_{i\in[m]}\E{\ell(z_i,z;\theta)}{z \sim p_{\mathcal{Z}}}\Big\vert &= 
\Big\vert \E{\hat{\mathcal{L}}(z;\theta)}{z \sim p_\mathcal{Z}} - \tfrac{1}{m} \textstyle\sum\limits_{i\in[m]} \hat{\mathcal{L}}(z_i;\theta) \Big\vert \\
&\leqslant \zeta\,\delta_\mathcal{S} + L\,\sqrt{\dfrac{2\,K\,\log 2+2\,\log\nicefrac{1}{\gamma}}{m}}
\end{split}
\end{equation}
Note that  $K$ is dependent on $\delta_s$ and satisfies $\displaystyle\lim_{m\to\infty}\tfrac{K}{m}\to 0$ ensuring that the right hand side goes to zero as more samples are exploited and the covering radius is improved. Thus, asymptotically the following holds:
$$\Big\vert\E{\ell(z_i,z_j;\theta)}{z_i,z_j} - \tfrac{1}{m}\!\textstyle\sum\limits_{i\in[m]}\E{\ell(z_i,z_j;\theta)}{z_j}\Big\vert \leqslant \mathcal{O}(\delta_s) + \mathcal{O}(\sqrt{\nicefrac{\log\tfrac{1}{\gamma}}{m}}) \text{ with probability at least }1-\gamma\,. $$
\end{proof}

\subsection{Proof for Proposition 4.2}
\begin{customprop}{4.2}\label{prop42}
Given $\lbrace z_i\rbrace_{i\in[n]} \overset{i.i.d.}{\sim} p_\mathcal{Z}$ and a set $s\subset[n]$. If $s=\cup_k s_k^\prime$ with $s_k^\prime$ is the $\delta_s$-cover of $\lbrace i\in [n] \mid y_i=k\rbrace$ (\ie, the samples in class $k$ ), $\ell(z_i,z_j;\theta)$ is $\zeta$-Lipschitz in $x_i, x_j$ for all $y_i$, $y_j$ and $\theta$, and bounded by $L$,  $e(\mathcal{A}_{s\sxtimes[n]})$ training error; then with probability at least $1-\gamma$ we have: 
$$ \Big\vert \tfrac{1}{n^2}\!\!\!\!\!\!\!\textstyle\sum\limits_{i,j\in[n]\sxtimes[n]} \!\!\!\!\!\!\! \ell(z_i,z_j;\mathcal{A}_{s\sxtimes[n]}) - 
\tfrac{1}{\vert s \vert\,n}\!\!\!\!\!\!\!\textstyle\sum\limits_{i,j\in s\sxtimes[n]}\!\!\!\!\!\! \ell(z_i,z_j;\mathcal{A}_{s\sxtimes[n]}) \Big\vert 
\leqslant
\mathcal{O}(\zeta\,\delta_s) + \mathcal{O}(e(\mathcal{A}_{s\sxtimes[n]})) + \mathcal{O}(L\,\sqrt{\nicefrac{\log\tfrac{1}{\gamma}}{n}})$$
\end{customprop}

\begin{proof}

We are given a condition on $s$ that we can partition $\mathcal{Z}$ into $m=\vert s\vert$ disjoint sets such that any sample from the dataset $(x_i, c), i\in[n]$, has a corresponding sample from $s$, $(x_j^\prime, c), j \in s$ within $\delta_s$ ball. Thus, we start with partitioning $\mathcal{Z}$ into $s$ disjoint sets as $\mathcal{Z}=\cup_i\mathcal{S}_i$ with $\mathcal{S}_i\cap\mathcal{S}_j=\emptyset,\,\forall i\neq j$.

We define $\ell_{[n]}(z)=\tfrac{1}{n}\sum\limits_{i\in[n]}\ell(z, z_i, \mathcal{A}_{s\sxtimes[n]})$ and $\ell_{s}(z)=\tfrac{1}{m}\sum\limits_{i\in s}\ell(z, z_i, \mathcal{A}_{s\sxtimes[n]})$ for the sake of clarity. Hence, we are interested in bounding $\vert \tfrac{1}{n}\sum_{[n]}\ell_{[n]}(z_i) -\tfrac{1}{m}\sum_{s}\ell_{[n]}(z_i) \vert$. We proceed with using triangle inequality to write:
\begin{equation}
\nonumber
\begin{split}
\Big\vert \tfrac{1}{n}\textstyle\sum\limits_{i\in [n]}\ell_{[n]}(z_i) &-\tfrac{1}{m}\textstyle\sum\limits_{i\in s}\ell_{[n]}(z_i) \Big\vert \\ 
&\leqslant
\Big\vert \tfrac{1}{n}\textstyle\sum\limits_{i\in [n]}\ell_{[n]}(z_i) -\textstyle\sum\limits_{i\in s}\tfrac{n_i}{n}\ell_{[n]}(z_i) \Big\vert^{(T1)}
+ \Big\vert \textstyle\sum\limits_{i\in s}\tfrac{n_i}{n}\ell_{[n]}(z_i) -\tfrac{1}{m}\textstyle\sum\limits_{i\in s}\ell_{[n]}(z_i) \Big\vert^{(T2)}
\end{split}
\end{equation}
For term $(T1)$ we write:
$$ (T1) \leqslant \tfrac{1}{n}\textstyle\sum\limits_{i\in [m]}\sum\limits_{z_j\in \mathcal{S}_i}\vert\ell_{[n]}(z_{s(i)}) -\ell_{[n]}(z_j)\vert
\overset{(1)}{\leqslant} \zeta\,\delta_s$$
where in $(1)$, we use $\zeta$-Lipschitz of the loss function and the condition $\vert z_{s(i)} - z_j\vert\leqslant \delta_s,\, \forall z_j\in \mathcal{S}_i$.

Using triangle inequality, we bound term $(T2)$ as:
\begin{equation}
\nonumber
\begin{split}
\Big\vert \textstyle\sum\limits_{i\in s}\tfrac{n_i}{n}\ell_{[n]}(z_i) & -\tfrac{1}{m}\textstyle\sum\limits_{i\in s}\ell_{[n]}(z_i) \Big\vert 
\leqslant
\Big\vert \E{\ell_{s}(z)}{z\sim p_\mathcal{Z}} - \E{\ell_{[n]}(z)}{z\sim p_\mathcal{Z}} \Big\vert^{(T2.1)} \\
&+
\Big\vert \E{\ell_{[n]}(z)}{z\sim p_\mathcal{Z}} - \textstyle\sum\limits_{i\in s}\tfrac{n_i}{n}\ell_{[n]}(z_i) \Big\vert^{(T2.2)}
+
\Big\vert \E{\ell_{s}(z)}{z\sim p_\mathcal{Z}} - \tfrac{1}{n}\textstyle\sum\limits_{i\in [n]}\ell_{s}(z_i) \Big\vert^{(T2.3)}
\end{split}
\end{equation}
where we use $\tfrac{1}{m}\textstyle\sum_{s}\ell_{[n]}(z_i) = \tfrac{1}{n}\textstyle\sum_{[n]}\ell_{s}(z_i)$ in $(T2.3)$.

For $(T2.1)$ we have:
$$(T2.1)\leqslant
\Big\vert \E{\tfrac{1}{m}\textstyle\sum\limits_{i\in s}\ell(z_i, z) - \tfrac{1}{n}\textstyle\sum\limits_{i\in [n]}\ell(z_i, z)}{z\sim p_\mathcal{Z}}\Big\vert$$
where we abuse the notation for the sake of clarity and drop parameter $\mathcal{A}_{s\sxtimes[n]}$ dependency from the loss. Rearranging the terms, we have:
$$(T2.1)\leqslant
\Big\vert \E{\tfrac{1}{m}\textstyle\sum\limits_{i\in [m]}
\tfrac{n-m\,n_i}{n}\ell(z_{s(i)}, z)}{z\sim p_\mathcal{Z}}\Big\vert
+\Big\vert \E{\tfrac{1}{n}\textstyle\sum\limits_{i\in [m]}\textstyle\sum\limits_{j\in \mathcal{S}_i}
\ell(z_{s(i)}, z) - \ell(z_j, z)}{z\sim p_\mathcal{Z}}\Big\vert$$
where similar to $(T1)$, the second summand is upper bounded by $\zeta\,\delta_s$. Using triangle inequality for the first summand, we write:
$$\Big\vert \E{\tfrac{1}{m}\textstyle\sum\limits_{i\in [m]}
\tfrac{n-m\,n_i}{n}\ell(z_{s(i)}, z)}{z\sim p_\mathcal{Z}}\Big\vert \leqslant (T2.3) + e(\mathcal{A}_{s\sxtimes[n]}) $$
Hence, we have:
$$ (T2.1)\leqslant \zeta\,\delta_s +  (T2.3) + e(\mathcal{A}_{s\sxtimes[n]})$$
where from Hoeffding’s Bound, $ (T2.3) \leqslant L\sqrt{\nicefrac{\log\tfrac{1}{\gamma}}{2n}}$ with probability at least $1-\gamma$:

Finally, we express $(T2.2)$ as:
\begin{equation}
\nonumber
\begin{split}
(T2.2) &= \Big\vert \textstyle\sum\limits_{i\in [m]}\E{\ell_{[n]}(z)\mid z\in\mathcal{S}_i}{z\sim p_\mathcal{Z}}p(z\in\mathcal{S}_i) - \textstyle\sum\limits_{i\in s}\tfrac{n_i}{n}\ell_{[n]}(z_i) \Big\vert \\
&\leqslant 
\Big\vert \textstyle\sum\limits_{i\in [m]}\E{\ell_{[n]}(z)\mid z\in\mathcal{S}_i}{z\sim p_\mathcal{Z}}\tfrac{n_i}{n} - \textstyle\sum\limits_{i\in s}\tfrac{n_i}{n}\ell_{[n]}(z_i)\Big\vert \\
&+
\Big\vert \textstyle\sum\limits_{i\in [m]}\E{\ell_{[n]}(z)\mid z\in\mathcal{S}_i}{z\sim p_\mathcal{Z}}p(z\in\mathcal{S}_i) - \textstyle\sum\limits_{i\in [m]}\E{\ell_{[n]}(z)\mid z\in\mathcal{S}_i}{z\sim p_\mathcal{Z}}\tfrac{n_i}{n}\Big\vert
\end{split}
\end{equation}
Rearranging the terms we have:
$$(T2.2) \leqslant 
 \textstyle\sum\limits_{i\in [m]} \tfrac{n_i}{n} \max_{z\in\mathcal{S}_i}\vert\ell_{[n]}(z) - \ell_{[n]}(z_{s(i)})\vert
+
\max_{z\in\mathcal{Z}}\vert\ell_{[n]}(z)\vert\textstyle\sum\limits_{i\in [m]}\Big\vert\tfrac{n_i}{n} - p(z\in\mathcal{S}_i)\Big\vert $$
where the first summand is bounded above by $ \zeta\, (\delta_s + \varepsilon(n))$ owing to loss being $\zeta$-Lipschitz. Here, we denote $\varepsilon(n)$ as the covering radius of $\mathcal{Z}$, \ie, the dataset, $\lbrace x_i, y_i \rbrace_{[n]}$ is  $\varepsilon(n)$-cover of $\mathcal{X}\sxtimes\mathcal{Y}$. We note that $(n_i)_{i\in[m]}$ is an $i.i.d.$ multinomial random variable with parameters $n$ and $(p_{\mathcal{Z}}(z\in\mathcal{S}_i))_{i\in[m]}$. Thus, by the Breteganolle-Huber-Carol inequality (Proposition A6.6 of \cite{van1996weak}), we have :
$$(T2.2)\leqslant \zeta\, (\delta_s + \varepsilon(n)) + L\,\sqrt{\tfrac{2m\log2+2\log{\nicefrac{1}{\gamma}}}{n}}$$
Finally, with probability at least $1-\gamma$, we end up with:
$$\Big\vert \tfrac{1}{n}\textstyle\sum\limits_{i\in [n]}\ell_{[n]}(z_i) -\tfrac{1}{m}\textstyle\sum\limits_{i\in s}\ell_{[n]}(z_i) \Big\vert
\leqslant \zeta\,(3\,\delta_s + \varepsilon(n)) + e(\mathcal{A}_{s\sxtimes[n]}) + L\,(\sqrt{\nicefrac{\log\tfrac{1}{\gamma}}{2n}} + \sqrt{\tfrac{2m\log2+2\log{\nicefrac{1}{\gamma}}}{n}}) $$

\end{proof}

\begin{customcorr}{4.2.1}\label{corr421}
Generalization performance of the proxy-based methods can be limited by the maximum of distances between the proxies and the corresponding class samples in the dataset.
\end{customcorr}

\begin{proof}
The covering radius for each class subset is the maximum distance between the corresponding class samples and the class proxy. We at least know that the generalization error is bounded above with a term proportional to that distance.
\end{proof}

\fi

\end{document}